\documentclass[11pt,a4paper]{deepseek}

\usepackage{amssymb}
\usepackage[numbers,sort&compress]{natbib}
\usepackage{multirow}
\usepackage{algorithm}
\usepackage{algpseudocode}
\usepackage{xspace}
\usepackage{placeins}
\usepackage{needspace}
\usepackage{adjustbox}
\usepackage{xcolor}
\usepackage{hyperref}

\definecolor{darkblue}{RGB}{0, 0, 139}
\hypersetup{
    colorlinks=true,
    urlcolor=darkblue
}

\graphicspath{{figures/}}

\newcommand{\ours}{\mbox{SE-JEPA}\xspace}

\newcommand{\memory}{\mathcal{M}}
\newcommand{\replay}{\mathcal{R}}
\newcommand{\loss}{\mathcal{L}}

\DeclareMathOperator*{\argmin}{arg\,min}

\newcommand{\missing}{\textemdash}

\newcommand{\supref}[1]{Appendix~#1}


\makeatletter
\DeclareRobustCommand\onedot{\futurelet\@let@token\@onedot}
\def\@onedot{\ifx\@let@token.\else.\null\fi\xspace}
\def\eg{\emph{e.g}\onedot}

\makeatother

\DeclareRobustCommand{\paperxref}[3]{\mbox{#1~\ref{#2}#3}}
\DeclareRobustCommand{\appendixref}[1]{\mbox{Appendix~\ref{#1}}}
\newcommand{\maincite}[2]{\cite{#1}}

\title{SE-JEPA: Self-Evolving Latent World Models}

\makeatletter
\def\@author{%
  \begin{tabular}{@{}c@{}}
    Chen Lin\textsuperscript{1,*}, Tu Yiyang\textsuperscript{2,*}, Jiacheng Zhou\textsuperscript{3}, Haotian Pu\textsuperscript{3},\\[2pt]
    Bin Li\textsuperscript{4}, Longtai Zhang\textsuperscript{3}, Shuangkang Fang\textsuperscript{5,\dag}, Hongwei Wang\textsuperscript{1,\dag}
  \end{tabular}%
}
\makeatother
\hypersetup{pdfauthor={Chen Lin, Tu Yiyang, Jiacheng Zhou, Haotian Pu, Bin Li, Longtai Zhang, Shuangkang Fang, Hongwei Wang}}

\correspondingauthor{*\ Equal contribution\qquad\dag\ Corresponding authors}

\newcommand{\affillogo}[4][0.18]{%
    \begin{minipage}[t]{#1\textwidth}
        \centering
        \includegraphics[height=1.1cm]{#2}\\[3pt]
        {\footnotesize\color{mornavy}\textsuperscript{#3}#4\par}
    \end{minipage}}
\institutionlogos{%
    \affillogo{logos/tongji.png}{1}{Tongji University}\hfill
    \affillogo{logos/zjut.pdf}{2}{Zhejiang University\\of Technology}\hfill
    \affillogo{logos/fudan.png}{3}{Fudan University}\hfill
    \affillogo[0.22]{logos/siat.png}{4}{Shenzhen Institutes of\\Advanced Technology}\hfill
    \affillogo{logos/buaa.png}{5}{Beihang University}%
}

\reportname{SE-JEPA}
\version{arXiv preprint}


\begin{document}
\sloppy
\maketitle

\begin{abstract}
\looseness=-1
We propose \ours{}, a framework that turns deployment experience into reusable predictive knowledge, enabling world models to learn continually under recurring environmental changes. 
The method decouples fast within-episode adaptation from cross-episode learning, building predictive capabilities by replaying observed interaction trajectories rather than directly inheriting episode-specific weight updates. 
\ours{} automatically groups predictively compatible trajectories into predictive domains and trains a separate slow predictor for each domain in a fixed representation space. Prediction errors guide experience selection and predictor validation, while current interactions determine which predictor to use for planning. 
The framework requires neither predefined domain labels nor domain-switch signals and supports both gradient-based and sampling-based planners. 
Experiments show that \ours{} improves test success over AdaJEPA by 15.4--17.1 percentage points on unseen PushObj shapes during pretraining and 15.9--18.7 points on PushT visual shifts. In a repeated-goal stream spanning 14 shape and visual conditions, online success increases from 65.3\% to 75.5\% over five deployment rounds. 
Furthermore, with DINO-WM as the backbone, a predictor trained through a single consolidation pass reduces Chamfer distance by 47\% on rope instances excluded from learning, without within-episode adaptation. These results demonstrate that \ours{} not only corrects current predictions but also enables world models to continually improve subsequent planning.
\vspace{1mm}
\\
\textcolor{darkblue}{\textbf{Website}}: \href{https://celveen.github.io/SE-JEPA-Page/}{https://celveen.github.io/SE-JEPA-Page/} \\
\textcolor{darkblue}{\textbf{Code}}: \href{https://github.com/Celveen/SE-JEPA}{https://github.com/Celveen/SE-JEPA}

\end{abstract}

\section{Introduction}
\label{sec:intro}

Latent world models provide an efficient approach to planning from visual observations by predicting action-conditioned state evolution in representation space~\cite{lecun2022path,zhou2024dinowm,assran2025vjepa2,maes2026lewm}. However, their performance depends on how well the learned model matches the deployment environment. When object shapes, visual conditions, or dynamics deviate from the training distribution, model predictions may diverge from the actual state evolution, causing planning to rely on inaccurate future states.
To address this challenge, 
AdaJEPA~\cite{wang2026adajepa} adapts its model after each executed action block but resets these adaptations between episodes.
JEPA-TTT~\cite{zhang2026jepattt} extends predictor learning across episodes by accumulating interactions and replaying them. Yet continued updating does not, by itself, yield reusable knowledge. 
Our diagnostic experiments highlight the limitations of simply retaining adapted weights: 
On PushObj, inheriting adapted weights reduces success from 62\% to 48\%. In a separate deployment stream, consecutive fast-update fingerprints show weak directional alignment (cosine similarity: 0.018). 
These observations suggest that parameter corrections fitted to individual interaction segments need not benefit subsequent trajectories. Under recurring conditions, deciding which experiences should be learned jointly and when a learned predictor should be reused therefore remains a central challenge for test-time adaptation.

\begin{figure*}[t]
  \centering
  \includegraphics[width=\textwidth]{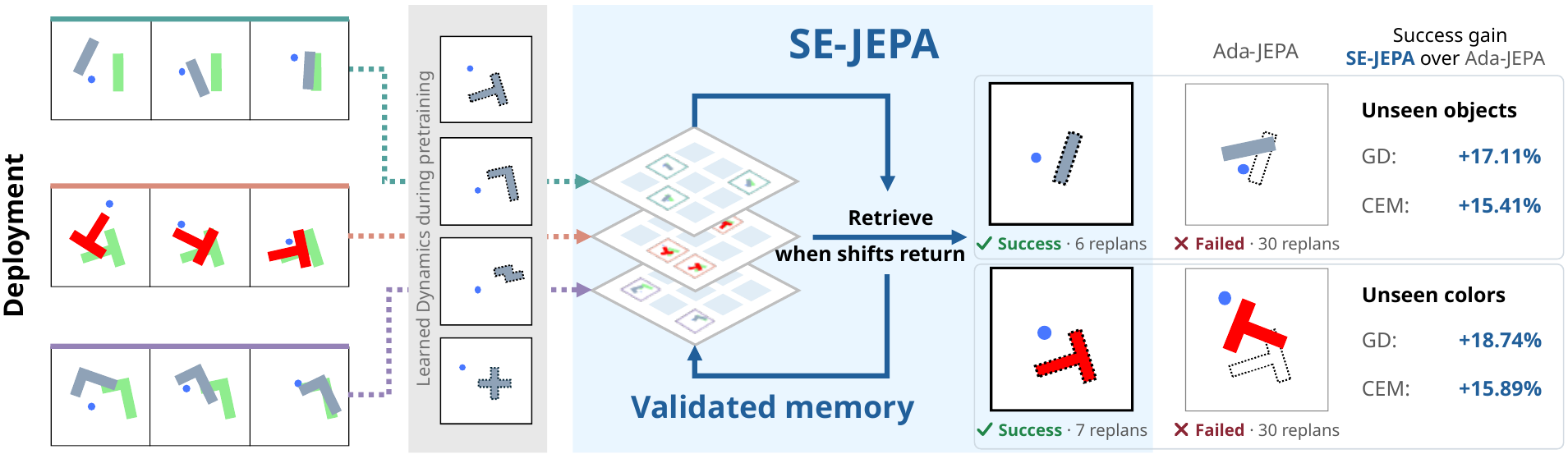}
  \captionof{figure}{\textbf{From episodic adaptation to reusable predictive knowledge}. Rather than carrying episode-specific parameter updates forward, SE-JEPA consolidates interaction trajectories into persistent, validated memory and selectively retrieves compatible predictors when conditions recur. This enables world models to leverage past experience for future planning, achieving higher success than AdaJEPA on unseen object shapes and color shifts under both gradient-descent (GD) and cross-entropy method (CEM) planning.}
  \label{fig:teaser}
\end{figure*}

\looseness=-1
Building on the above insight, we introduce \ours{}, a latent world model framework that enables continual learning across episodes through persistent predictive memory. Inspired by complementary learning systems~\cite{mcclelland1995cls}, \ours{} separates fast within-episode adaptation from slow cross-episode consolidation. The fast system adapts a temporary copy of the model to the current environment, maintaining responsiveness to local changes. The slow system replays historical interaction trajectories to progressively learn dynamics predictors that can be reused for future goals. Accumulating experience over time, however, also requires addressing compatibility: trajectories collected under different conditions may not be well modeled by a single predictor. To this end, \ours{} organizes past experience into self-discovered \emph{predictive domains}. Predictively compatible trajectories are grouped into the same domain, each with its own replay buffer and slow predictor. This allows dynamics knowledge from different environments to accumulate separately, rather than forcing it into a single set of parameters.

Furthermore, \ours{} uses its own prediction errors to select which experiences to retain and when to reuse learned predictors. Trajectories that the pretrained model already predicts accurately need not be written directly to memory, whereas those with substantial prediction errors support continued learning within the corresponding predictive domains. A slow predictor becomes eligible for retrieval only after consistently outperforming the pretrained model on new trajectories evaluated before it is trained on them. Thus, in \ours{}, historical trajectories establish whether learned knowledge is eligible for reuse, while current interactions determine whether that knowledge is relevant to the present planning problem. This mechanism links experience accumulation to selective reuse, rather than unconditionally applying past adaptations to new interactions.

Our main contributions are summarized as follows:
\begin{itemize}
    \item 
    \looseness=-1
    We introduce \ours{}, which separates fast within-episode adaptation from slow cross-episode consolidation, learning reusable predictors from deployment trajectories rather than accumulating \mbox{episode-specific weight updates}.

    \item 
    We develop a label-free memory that organizes compatible trajectories into predictive domains. Prediction errors guide experience selection, validate predictors on new trajectories before learning from them, and determine which predictor to reuse for current planning.

    \item Extensive experiments demonstrate the effectiveness of \ours{} in cross-episode knowledge reuse, improved planning under distribution shifts, and continual performance gains during deployment.

\end{itemize}

\begin{figure*}[t]
  \centering
  \includegraphics[width=\textwidth]{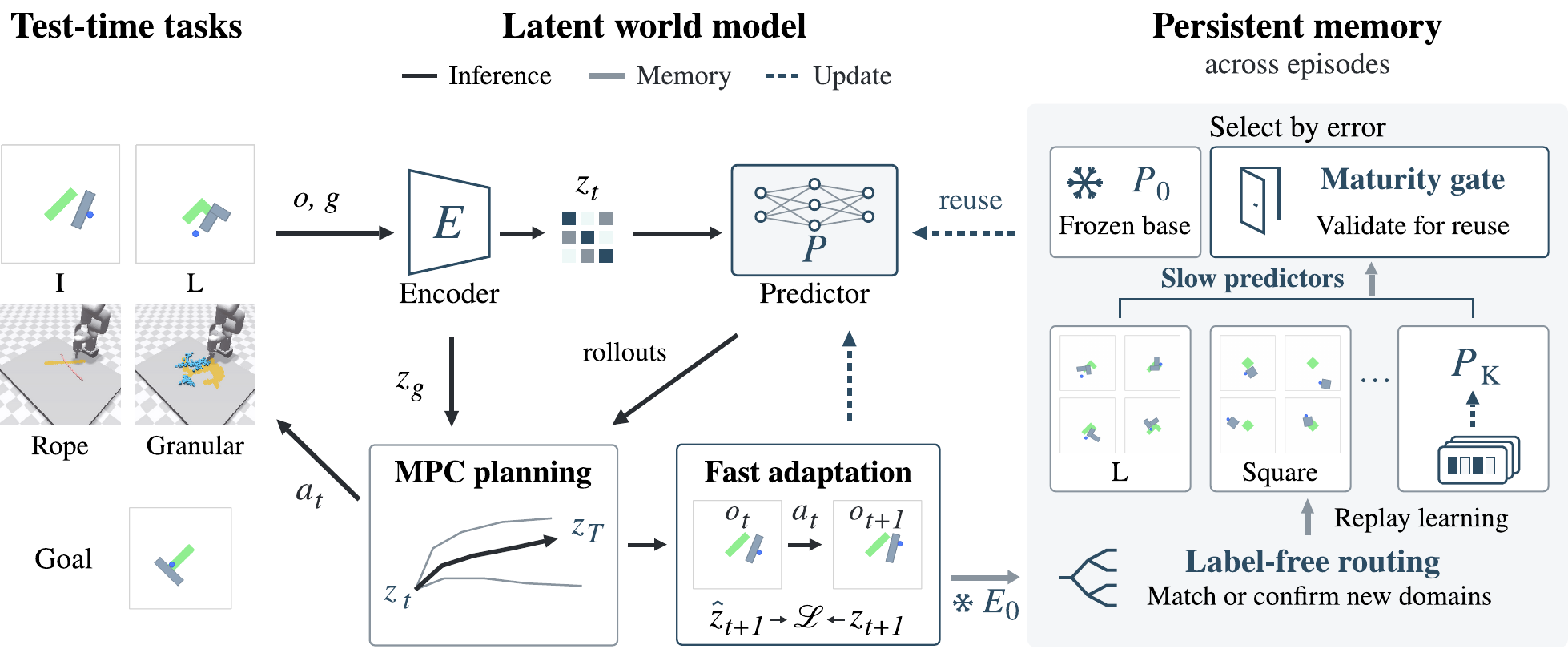}
  \caption{\textbf{Overview of \ours.} 
Given test-time observations and goals (\textit{left}), the latent world model performs MPC planning and fast adaptation using observed transitions (\textit{middle}). After each episode, trajectories are re-encoded by the frozen encoder and organized into predictively compatible domains through label-free routing, where slow predictors learn reusable dynamics via replay (\textit{right}). A maturity gate validates slow predictors on new trajectories before allowing their reuse. During subsequent episodes, prediction errors from early interactions determine whether to retrieve a validated predictor or retain the frozen base for planning. Fast adaptations reset after each episode, while predictive memory persists across episodes.}
  \label{fig:architecture}
\end{figure*}

\section{Related Work}
\label{sec:related}

\noindent \textbf{World models and representation learning}.
World models enable planning with learned dynamics~\cite{watter2015e2c,chua2018pets,hafner2019planet}, policy learning through imagined rollouts~\cite{ha2018worldmodels,kaiser2020simple,micheli2023iris,hafner2020dreamer,hafner2021dreamerv2,hafner2025dreamerv3}, tree search~\cite{schrittwieser2020muzero,ye2021efficientzero}, and value-guided latent control~\cite{hansen2022tdmpc,hansen2024tdmpc2}.
In parallel, predictive representation learning~\cite{lecun2022path,oord2018cpc,schwarzer2021spr}, non-contrastive self-supervision~\cite{grill2020byol,zbontar2021barlow,bardes2022vicreg}, and visual representation models~\cite{caron2021dino,oquab2024dinov2,nair2022r3m} have enabled learning in increasingly expressive latent spaces.
Building on these advances, image and video JEPA~\cite{assran2023ijepa,bardes2024vjepa,assran2025vjepa2} and action-conditioned latent world models~\cite{zhou2024dinowm,maes2026lewm,wang2026straightening,yin2026ddpwm} have advanced representation learning and predictive planning.
While these methods focus primarily on learning representations and dynamics, \ours{} investigates how deployment experience can be consolidated into reusable predictive knowledge across recurring distribution shifts.

\vspace{1mm}
\noindent \textbf{Test-time adaptation.}
Test-time training and entropy minimization update deployed models from unlabeled observations~\cite{sun2020ttt,wang2021tent}, including under changing distributions~\cite{wang2022cotta}. In control, each executed action supplies a target for the dynamics predictor. AdaJEPA~\cite{wang2026adajepa} uses this signal within an episode and resets the weights at its end. JEPA-TTT~\cite{zhang2026jepattt} studies a dynamics shift that persists across episodes. It trains one predictor by replaying overlapping windows from the accumulated interactions. \ours addresses recurring shifts. It keeps the pretrained weights fixed, groups compatible trajectories, and retrieves separate slow predictors whose errors on new trajectories are lower than those of the pretrained reference. These comparisons precede replay training on each trajectory.

\vspace{1mm}
\noindent\textbf{Continual learning and complementary memory.}
Complementary learning systems distinguish rapid adaptation from gradual knowledge consolidation~\cite{mcclelland1995cls}.
Continual learning mitigates forgetting through experience replay~\cite{rolnick2019replay}, constrained updates~\cite{lopezpaz2017gem,kirkpatrick2017ewc}, knowledge distillation~\cite{li2016lwf,fang2026pvdal,fang2023pvd}, and modular architectures~\cite{rusu2016progressive}.
In world model planning, however, reusing a predictor directly influences subsequent actions and observations, making knowledge selection as important as retention.
\section{Method}
\label{sec:method}

\subsection{Motivation}

During deployment, a world model repeatedly plans for new goals and observes the outcomes of its actions.
Within-episode adaptation uses this feedback to improve current predictions, but the resulting updates are discarded when the model resets.
A natural alternative is to retain the adapted weights for subsequent episodes.
However, updates fitted to recent interactions may not generalize to the states and actions encountered under a different goal.
Our experiments reveal this limitation.
On 100 matched PushObj goals, directly inheriting fast weights reduces success from 62\% to 48\% compared with episodic reset (Table~\ref{tab:inheritance}).
In a separate seven-round stream, consecutive fast-update directions show little alignment, with a mean cosine similarity of 0.018 under both GD and CEM (see \supref{}).
These results motivate learning from the underlying interaction data rather than directly accumulating episode-specific updates.

\subsection{\ours{} Overview}
Observed trajectories record how the environment responds to actions, even when a task fails.
Replaying these trajectories allows a predictor to learn from varied contacts and motions across episodes, rather than only recent interactions.
As shown in Fig.~\ref{fig:architecture}, \ours{} turns this experience into persistent predictive memory while retaining fast within-episode adaptation.
Each episode starts from working copies of the frozen pretrained encoder--predictor pair $(E_0, P_0)$ and alternates MPC planning with fast adaptation on observed transitions.
During early replans, the frozen reference assesses familiarity through prediction errors.
Familiar episodes retain $P_0$ as their adaptation base; otherwise, the lowest-error predictor among $P_0$ and eligible slow predictors becomes the new base.
At episode end, the trajectory is re-encoded with $E_0$ to keep cross-episode learning in a fixed latent space.
For unfamiliar trajectories, retrieval results or fast-update fingerprints propose candidate domains, and prediction errors determine whether to accept the trajectory.
Unmatched experience remains pending until sufficient evidence supports a new domain.
Each domain maintains a separate replay buffer and a slow predictor initialized from $P_0$.
The predictor learns from accumulated trajectories through replay.
Before training on a new trajectory, it is evaluated against $P_0$; consistent predictive improvement makes it eligible for future retrieval.
Fast-adaptation parameters reset between episodes, while slow predictors, replay buffers, and validation records persist.
This separation allows experience to accumulate without directly carrying episode-specific updates into subsequent tasks.

\subsection{Fast Adaptation and Replay Consolidation}
\label{sec:fast}
\label{sec:slow}

Within each episode, MPC optimizes an action sequence to minimize the latent goal cost and executes its first block.
After observing the outcome, the model adapts using prediction errors on recent transitions.
Following AdaJEPA~\cite{wang2026adajepa}, we take one Adam step to update the predictor's last block and the encoder head before replanning.
These updates affect only the episode-local working copies.

For cross-episode learning, each predictive domain $k$ maintains a replay buffer $\replay_k$ and a slow predictor $P_{\phi_k}$, where $\phi_k$ denotes its parameters.
Accepted trajectories are re-encoded with the frozen encoder $E_0$, so replay does not mix representations from different fast encoder states.
Each slow predictor is initialized from $P_0$ and trained on sampled replay windows:
\begin{equation}
  \loss_k(\phi_k)=
  \mathbb{E}_{(x,z^+)\sim\replay_k}
  \big\|P_{\phi_k}(x)-z^+\big\|_2^2,
  \label{eq:replay}
\end{equation}
where $x$ contains a short history of observation embeddings and actions, and $z^+$ is the next observation embedding.

\subsection{Discovering Predictive Domains}
\label{sec:write}

We first assess whether the frozen reference already predicts an episode accurately.
Let $\ell^{(0)}_{e,t}$ denote its next-observation prediction error at transition $t$ of episode $e$.
We compute the geometric mean over up to the first three observed transitions, indexed by $\mathcal{I}_e$:
\begin{equation}
  F_e=\exp\!\left(
  \frac{1}{|\mathcal{I}_e|}
  \sum_{t\in\mathcal{I}_e}\log\ell^{(0)}_{e,t}
  \right).
  \label{eq:familiarity}
\end{equation}
An episode is considered familiar if $F_e<\tau_f$, where $\tau_f$ is a calibrated familiarity threshold.
Such episodes retain $P_0$ as their adaptation base and skip direct memory writing.

For an unfamiliar episode, we concatenate the fast predictor and encoder-head updates into $\Delta_e$.
Its normalized direction serves as an update fingerprint:
\begin{equation}
  u_e=\frac{\Delta_e}{\|\Delta_e\|_2},
  \qquad s(e,k)=u_e^\top c_k,
  \label{eq:fingerprint}
\end{equation}
where $c_k$ is the unit-length fingerprint prototype of domain $k$, and $s(e,k)$ measures their cosine similarity.
If a slow predictor was retrieved during the episode, its domain is considered first; otherwise, candidates are ranked by $s(e,k)$.

Candidate domains are checked against the observed trajectory $\tau_e$ before any replay update.
We roll out each candidate predictor and $P_0$ under the same recorded actions and compare their errors:
\begin{equation}
  r_{e,k}=
  \frac{\loss_{\mathrm{roll}}(\phi_k;\tau_e)}
       {\loss_{\mathrm{roll}}(\theta_0;\tau_e)}.
  \label{eq:ratio}
\end{equation}
Here $\theta_0$ denotes the parameters of $P_0$, and $\loss_{\mathrm{roll}}$ measures open-loop prediction error in the fixed latent space defined by $E_0$.
In the rigid-object experiments, a candidate accepts the trajectory if $r_{e,k}\leq1.2$.

Unmatched trajectories enter a pending pool.
When an incoming fingerprint aligns with two pending fingerprints, the three trajectories form a new domain.
Existing domains merge when their predictors pass compatibility checks on each other's trajectories.
The larger domain retains its predictor and trains on the combined replay buffers.
Mature domains can also absorb smaller compatible groups (\supref{A}).

\begin{table*}[!t]
  \centering
  \caption{\textbf{Comparison with baselines on goal-conditioned planning with CEM.}
Success rates (\%) cover seven PushObj shapes (\emph{All-7}), including three unseen during pretraining (\emph{Unseen}), and six PushT visual shifts (\emph{All-6}), including three recolorings (\emph{Color-3}).
\emph{Replans} denotes the mean number of MPC replans, with five actions per replan and failed episodes assigned the maximum of 30 replans. CD is the final Chamfer distance to the goal.}
  \label{tab:main}
  \small
  \setlength{\tabcolsep}{3.5pt}
  \adjustbox{max width=\linewidth}{%
    \begin{tabular}{@{}lcccccccc@{}}
      \toprule
      & \multicolumn{3}{c}{PushObj}
      & \multicolumn{3}{c}{PushT visual shifts}
      & Rope & Granular\\
      \cmidrule(lr){2-4}
      \cmidrule(lr){5-7}
      \cmidrule(lr){8-8}
      \cmidrule(lr){9-9}
      Method & All-7 $\uparrow$ & Unseen $\uparrow$ & Replans $\downarrow$
      & All-6 $\uparrow$ & Color-3 $\uparrow$ & Replans $\downarrow$
      & CD $\downarrow$ & CD $\downarrow$\\
      \midrule
      DreamerV3~\cite{hafner2025dreamerv3} & 7.1 & 5.3 & 28.36 & 5.7 & 0.0 & 28.91 & 2.49 & 1.05 \\
IRIS~\cite{micheli2023iris} & 2.9 & 2.7 & 29.37 & 6.0 & 6.7 & 28.85 & 1.11 & 0.37 \\
R3M~\cite{nair2022r3m} & 18.9 & 18.0 & 25.49 & 21.7 & 27.3 & 25.00 & 1.13 & 0.95 \\
DINO CLS~\cite{oquab2024dinov2} & 14.9 & 10.0 & 26.77 & 34.7 & 35.3 & 22.38 & 0.84 & 0.79 \\
DINO-WM~\cite{zhou2024dinowm} & 62.0 & 46.0 & 16.15 & 33.0 & 37.3 & 22.05 & 0.491 & 0.301 \\
LeWM~\cite{maes2026lewm} & 64.0 & 50.0 & 16.28 & 40.3 & 16.0 & 21.45 & 0.786 & 0.322 \\
DDP-WM~\cite{yin2026ddpwm} & 57.7 & 43.3 & 17.59 & 52.7 & 57.3 & 18.39 & 0.677 & 0.259 \\
\midrule
Frozen (pretrained) & 43.8 & 36.2 & 21.08 & 48.3 & 33.6 & 20.10 & 0.491 & 0.301 \\
AdaJEPA~\cite{wang2026adajepa} & 70.4 & 54.4 & 16.02 & 60.2 & 42.2 & 18.46 & 0.629 & 0.304 \\
\ours & \textbf{80.4} & \textbf{73.1} & \textbf{12.99} & \textbf{76.1} & \textbf{65.2} & \textbf{14.46} & \textbf{0.252} & \textbf{0.220} \\
 
      \bottomrule
    \end{tabular}%
  }
\end{table*}

\begin{figure*}[t]
  \centering
  \includegraphics[width=1.\textwidth]{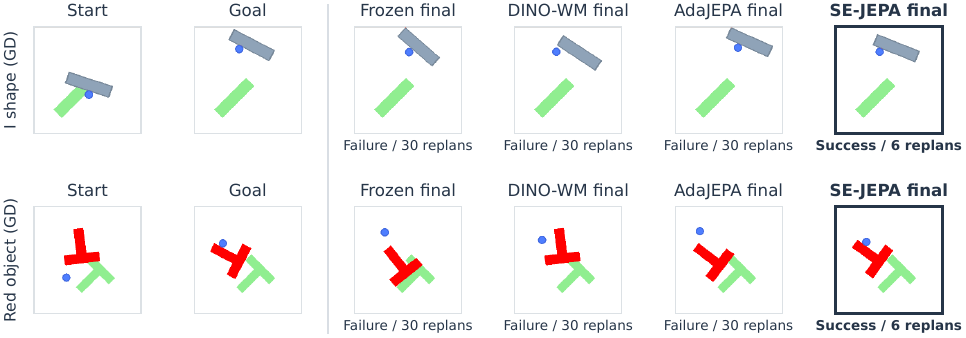}
  \caption{\textbf{Qualitative comparison with baselines.} For the unseen `I' shape (top) and mixed-stream red object (bottom), \ours{} succeeds in 6 replans in both cases, whereas baselines remain unsuccessful after 30 replans.}
  
  \label{fig:qualitative}
\end{figure*}

\begin{figure}[ht]
  \centering
  \includegraphics[width=\linewidth]{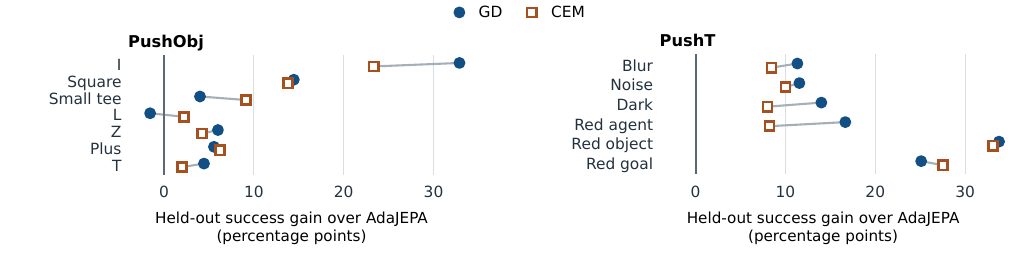}
  \caption{\textbf{Held-out success gains over AdaJEPA.} PushObj (left) and PushT (right); positive values favor \ours. `I', `square', and `small tee' are unseen shapes during pretraining.}
  \label{fig:condition_gains}
\end{figure}

\begin{figure*}[t]
  \centering
  \includegraphics[width=\textwidth]{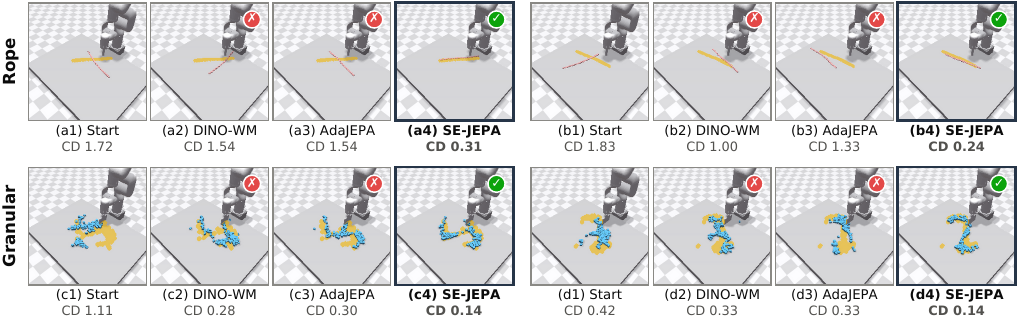}
  \caption{\textbf{Comparison on deformable manipulation task.} Each example shows the initial state and each method's final state after 15 pushes in FleX, for rope (top) and granular material (bottom). Gold indicates the target state. \ours{} outperforms both baselines.}
  \label{fig:deform_qualitative}
\end{figure*}

\begin{figure}[t]
  \centering
  \includegraphics[width=\linewidth]{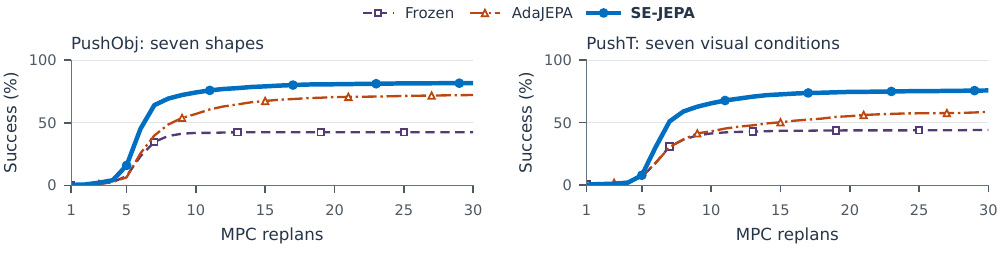}
\caption{\textbf{Planning efficiency versus baselines (GD).} Curves show success rates (\%) within each replan budget on all PushObj shapes (left) and PushT visual conditions (right). Test goals are excluded from memory learning; each replan executes five actions.}
  \label{fig:efficiency_main}
\end{figure}

\subsection{Validated Retrieval for Planning}
\label{sec:trust}

For each existing slow predictor, we record the pre-update ratios $r_{e,k}$ from accepted trajectories.
The predictor becomes eligible for retrieval when the arithmetic mean of its last three validation ratios is at most $0.9$.
This stricter threshold requires improvement over $P_0$ on new trajectories before the predictor can guide planning.

During the first three replans of an unfamiliar episode, we compare eligible slow predictors with $P_0$ on the transitions observed so far, denoted by $\mathcal{B}_{e,t}$.
We select the predictor with the lowest error:
\begin{equation}
  k^\star=
  \argmin_{k\in\{0\}\cup\mathcal{K}_{\mathrm{eligible}}}
  \loss_{\mathrm{pred}}(\phi_k;\mathcal{B}_{e,t}),
  \qquad \phi_0=\theta_0.
  \label{eq:retrieval}
\end{equation}
Here $\mathcal{K}_{\mathrm{eligible}}$ contains the eligible domain indices, and $\loss_{\mathrm{pred}}$ measures prediction error on the observed transitions.
The selected weights become the fast predictor's adaptation base, while the fast corrections already accumulated within the current episode are retained.
\section{Experiments and Results}
\label{sec:experiments}

\subsection{Settings}
\label{sec:setup}

\noindent \textbf{Tasks and backbones.}
For rigid-object manipulation, We use the same task-trained ResNet world models as AdaJEPA~\cite{he2016resnet,wang2026adajepa}.
\emph{PushObj} tests shape shifts: pretraining covers T, L, Z, and plus shapes, while deployment additionally includes I, square, and small tee.
\emph{PushT} tests visual shifts: it keeps the T shape and introduces blur, noise, darkness, or recoloring of the agent, object, or goal.
For deformable manipulation, we evaluate \emph{rope} and \emph{granular} using DINO-WM with pretrained DINOv2 features~\cite{zhou2024dinowm,oquab2024dinov2,lin2020softgym}.

\vspace{1mm}
\noindent \textbf{Deployment and evaluation.}
The separate-task PushObj and PushT comparisons use three learned states per planner and five rounds, resampling 50 targets per condition per round. 
We also evaluate each learned memory on separate test-seed streams with memory writes disabled and retrieval and fast adaptation active. 
We report success and mean replans, counting the full budget for failures. 
Rope and granular isolate slow consolidation: one predictor per task revisits 50 fixed instances over four passes, with fast adaptation disabled. We measure final Chamfer distance (CD) to the goal and separately test rope on new instances.

\vspace{1mm}
\noindent \textbf{Baselines.}
Our baselines without test-time training include DreamerV3~\cite{hafner2025dreamerv3}, IRIS~\cite{micheli2023iris}, R3M~\cite{nair2022r3m}, DINO CLS~\cite{oquab2024dinov2}, DINO-WM~\cite{zhou2024dinowm}, LeWM~\cite{maes2026lewm}, and DDP-WM~\cite{yin2026ddpwm}.
For test-time training, we compare with AdaJEPA~\cite{wang2026adajepa}.
We also include Frozen, which uses the same pretrained initialization as \ours{} but keeps all model parameters fixed throughout testing.
All rigid-object planners share a 150-action budget.
The paired held-out comparisons evaluate Frozen, AdaJEPA, and \ours{} on the same targets.
More details can be found in the \supref{}.

\subsection{Comparison with Baseline Methods}
\label{sec:sota}

\noindent \textbf{Quantitative comparison.}
Table~\ref{tab:main} summarizes goal-conditioned planning performance with CEM.
On PushObj, \ours{} achieves fifth-round online success rates of 80.4\% across all seven shapes and 73.1\% on the three shapes absent from pretraining.
On PushT, success across 6 visual shifts increases from 60.2\% with AdaJEPA to 76.1\% with \ours{}.
The gain is larger on the three recoloring conditions, where success increases from 42.2\% to 65.2\%, an improvement of 23.0 percentage points.

To assess memory reuse on new goals, we compare both methods on identical test goals excluded from memory learning (Fig.~\ref{fig:condition_gains}).
During evaluation, \ours{} keeps its persistent memory fixed while retaining predictor retrieval and within-episode adaptation.
Under both GD and CEM, the largest gains occur on I and square shapes and on object and goal recoloring.
These results show that learned memory improves planning on new goals under recurring conditions, extending adaptation benefits beyond individual episodes.

\vspace{1mm}
\noindent \textbf{Qualitative results.}
Fig.~\ref{fig:qualitative} compares final states under GD from identical initial states and goals.
For the `I' shape absent from pretraining and the recolored PushT object, \ours{} reaches the goal in 6 replans, whereas Frozen, DINO-WM, and AdaJEPA remain unsuccessful after 30 replans.
For deformable manipulation (Fig.~\ref{fig:deform_qualitative}), \ours{} achieves lower final CD than DINO-WM and AdaJEPA in all four examples under a 15-push budget.
These improvements are obtained with fast adaptation disabled for \ours{}, illustrating that cross-episode consolidation can improve goal matching without within-episode model updates.

\vspace{1mm}
\noindent \textbf{Planning efficiency.}
Beyond individual examples, Fig.~\ref{fig:efficiency_main} evaluates interaction efficiency across the full test sets under GD.
Each curve reports the fraction of test goals reached within a given number of replans.
Across all seven PushObj shapes and seven PushT visual conditions, \ours{} exceeds AdaJEPA's 30-replan success rate within 15 replans, attaining higher success with half the interaction budget.

\begin{figure}[t]
  \centering
  \includegraphics[width=0.62\linewidth]{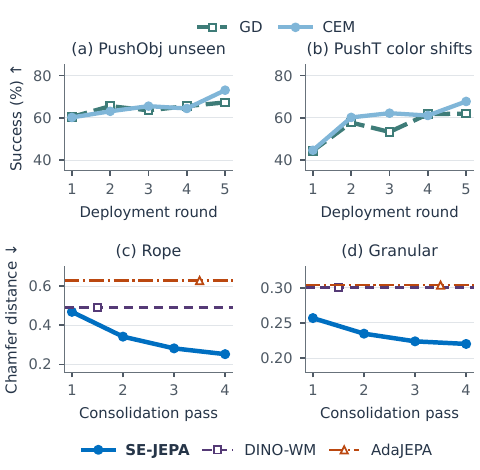}
  \caption{\textbf{Learning from deployment experience.}
(a,b): Online success on PushObj shapes unseen during pretraining and PushT color shifts, averaged over three runs with goals resampled each round.
(c,d): Mean final CD over four passes through 50 fixed instances per task, with fast adaptation disabled for \ours{}.}
  \label{fig:deployment}
\end{figure}

\begin{figure*}[t]
  \centering
  \includegraphics[width=\textwidth]{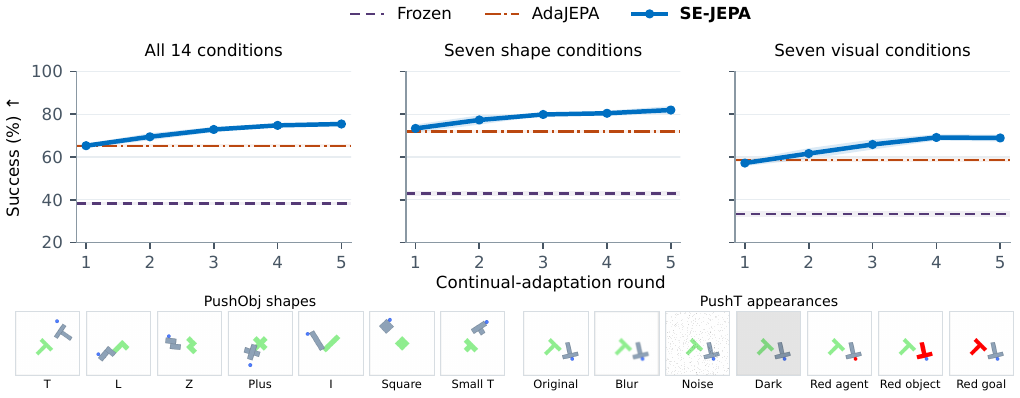}
\caption{\textbf{Learning across recurring conditions with a shared memory.} The curves show online success over five deployment rounds, with repeated initial--goal pairs and one persistent memory per stream. The bottom row illustrates the 14 shape and visual conditions.} 
  \label{fig:mixed14_learning}
\end{figure*}

\subsection{Learning from Deployment Experience}
\label{sec:deployment}
We examine whether deployment experience improves subsequent planning.
For ResNet-based world models, Fig.~\ref{fig:deployment}(a,b) tracks online success on unseen PushObj shapes and PushT color shifts.
Both GD and CEM achieve higher success in the fifth round than in the first, despite goals being resampled each round.
The gains therefore emerge across changing goals rather than repeated evaluation of fixed targets.
To isolate slow consolidation, we disable fast adaptation in DINO-WM-based \ours{} and revisit 50 fixed instances per task over four passes (Fig.~\ref{fig:deployment}(c,d)).
Mean final CD decreases across passes, reaching 0.252 on rope and 0.220 on granular material in the fourth pass, compared with 0.491 and 0.301 for frozen DINO-WM (Table~\ref{tab:main}).
These results show that cross-episode consolidation improves planning without within-episode parameter updates.
To assess generalization, we compare the predictor after one consolidation pass with pretrained DINO-WM on 91 rope instances excluded from learning.
With both predictors fixed throughout evaluation, \ours{} reduces mean final CD from 0.664 to 0.349, a 47\% reduction.
Thus, consolidation benefits extend to new instances without further predictor updates.

\subsection{Learning across Recurring Shifts}
\label{sec:mixed14}

We combine seven PushObj shapes and seven PushT visual conditions within each deployment stream to evaluate learning as conditions recur.
Each planner uses three learning streams, each spanning five rounds of 50 episodes per condition, with one memory retained throughout.
Initial/goal pairs repeat across rounds, while fast adaptation and memory updates remain active.
Averaged equally over GD and CEM, online success rises from 65.3\% to 75.5\% (Fig.~\ref{fig:mixed14_learning}).
Both task families improve: shapes from 73.4\% to 82.0\%, and visual conditions from 57.2\% to 68.9\%.
These gains show that cross-episode learning remains effective when shape and visual conditions recur within the same stream.

We further examine memory organization and retrieval across material changes.
A separate CEM stream alternates rope and granular tasks over three rounds, using a shared rope-pretrained DINO-WM and 50 new goals per material per round.
Fast adaptation and an additional absolute-error check for memory writes remain enabled.
Without material labels, \ours{} forms two predictive domains with 100\% purity among stored trajectories at each round endpoint (Table~\ref{tab:deform_mixed_main}).
Among slow-predictor selections, 98.9\% match the current material, reaching 100\% in the final two rounds.
Mean final CD decreases from 0.598 to 0.412 on rope and from 0.471 to 0.355 on granular material.
These results support label-free organization and selective reuse of material-specific experience, improving planning across recurring material changes.

\begin{table}[t]
  \centering
  \caption{\textbf{Learning across mixed rope and granular tasks (CEM).} Purity measures whether each learned domain contains trajectories from the same material. Retrieval accuracy measures how often the selected slow predictor matches the current material.}
  \label{tab:deform_mixed_main}
  \small
  \setlength{\tabcolsep}{5pt}
  \adjustbox{max width=\linewidth}{%
    \begin{tabular}{@{}lcccc@{}}
      \toprule
      Round & Rope CD $\downarrow$ & Granular CD $\downarrow$ & Purity (\%) $\uparrow$ & Retrieval (\%) $\uparrow$\\
      \midrule
      1 & 0.598 & 0.471 & 100 & 96.6 \\
2 & 0.452 & 0.369 & 100 & 100.0 \\
3 & 0.412 & 0.355 & 100 & 100.0 \\
 
      \bottomrule
    \end{tabular}%
  }
\end{table}

\subsection{Ablations Studies}
\label{sec:ablation}

\noindent \textbf{Parameter inheritance.}
We first examine whether directly retaining episode-specific parameter updates can improve subsequent planning.
We accumulate fast updates as $s_{e+1}=s_e+w\Delta_e$, where $s_e$ is the parameter offset from the pretrained model, $\Delta_e$ is the fast update from episode $e$, and $w$ controls the retained fraction.
On 100 matched PushObj goals (Table~\ref{tab:inheritance}), full inheritance ($w=1$) reduces success from 62\% with episodic reset ($w=0$) to 48\%, losing 25 previously successful goals while recovering only 11.
Although a small retention factor ($w=0.003$) slightly improves success to 64\%, larger values consistently degrade performance.
These results suggest that directly accumulating episode-specific corrections is unreliable for transferring knowledge across goals, motivating our trajectory-based consolidation.

\begin{table}[t]
  \centering
  \caption{\textbf{Fast-update inheritance versus episodic reset.} We evaluate planning success on 100 matched PushObj goals under different retention factors $w$.
Wins and losses count goals gained and lost relative to episodic reset ($w=0$), respectively.}
  \label{tab:inheritance}
  \small
  \setlength{\tabcolsep}{8pt}
  \begin{tabular}{@{}lcc@{}}
    \toprule
    Accumulation $w$ & Success (\%) & Wins / losses\\
    \midrule
    
$0$ (episodic reset) & 62.0 & \missing\\
$0.003$ & 64.0 & 7 / 5\\
$0.01$ & 59.0 & 5 / 8\\
$0.03$ & 54.0 & 4 / 12\\
$1$ (naive inheritance) & 48.0 & 11 / 25\\
 
    \bottomrule
  \end{tabular}
\end{table}

\vspace{1mm}
\noindent \textbf{Memory gates and slow learning rate.}
Table~\ref{tab:component_ablation} evaluates each variant after five learning rounds on 1,050 held-out goals, using one learning stream per task and planner.
All variants share the same task-specific initialization.
Removing familiarity protection improves PushObj success but reduces PushT success with both planners.
Removing maturity screening lowers PushT success with CEM but raises it with GD, indicating a planner-dependent effect.
Slow learning rates from $5\times10^{-5}$ to $3\times10^{-4}$ stay within 1.81 percentage points of the $10^{-4}$ default on PushObj and 3.52 on PushT.
The largest rate, $5\times10^{-4}$, reduces success in all four task--planner combinations.

\begin{table}[!ht]
  \centering
  \caption{\textbf{Ablation of memory gates and slow learning rates.} We report held-out success after five learning rounds on 1,050 goals across seven conditions per task.}
  \label{tab:component_ablation}
  \small
  \setlength{\tabcolsep}{8pt}
  \adjustbox{max width=\linewidth}{
    \begin{tabular}{@{}lcccc@{}}
      \toprule
      & \multicolumn{2}{c}{PushObj}
      & \multicolumn{2}{c}{PushT}\\
      \cmidrule(lr){2-3}\cmidrule(lr){4-5}
      Variant & GD & CEM & GD & CEM\\
      \midrule
      Full \ours & 81.43 & 78.57 & 76.57 & 77.90 \\
No maturity screening & 80.86 & 79.24 & 78.10 & 75.05 \\
No familiarity protection & 83.14 & 79.62 & 73.14 & 75.14 \\
\midrule
\multicolumn{5}{@{}l}{Full \ours: fixed slow learning rate}\\
$5\times10^{-5}$ & 79.62 & 79.14 & 73.05 & 75.81 \\
$10^{-4}$ (default) & 81.43 & 78.57 & 76.57 & 77.90 \\
$2\times10^{-4}$ & 82.38 & 78.10 & 76.86 & 76.76 \\
$3\times10^{-4}$ & 80.57 & 77.14 & 78.10 & 75.05 \\
$5\times10^{-4}$ & 78.38 & 74.38 & 74.38 & 73.52 \\
 
      \bottomrule
    \end{tabular}%
  }
\end{table}

\vspace{1mm}
\noindent \textbf{Prediction-guided retrieval.}
We further evaluate predictor selection on common CEM trajectories from 50 new PushT start--goal pairs under seven visual conditions.
The deployed rule uses the first three transitions to select among the pretrained reference and three mature predictors.
Prediction errors are measured only on subsequent observations, with all predictor weights fixed.
In Table~\ref{tab:mature_routing}, selection reduces one-step MSE by 77.6\% relative to the pretrained model and by 41.1\% relative to a calibrated fixed-domain baseline that also retains access to the pretrained model.
Five-step rollout error also decreases, showing that the selected predictors remain useful beyond the transitions used for selection.

\begin{table}[t]
  \centering
  \caption{\textbf{Error-guided predictor selection on unseen trajectories.} We report future one-step and five-step latent MSE ($\times10^{-3}$) across seven PushT visual conditions.
The oracle selects predictors using future prediction errors.}
  \label{tab:mature_routing}
  \small
  \setlength{\tabcolsep}{5pt}
   \adjustbox{max width=\linewidth}{
  \begin{tabular}{@{}lcc@{}}
    \toprule
    Predictor selection & One-step $\downarrow$ & Five-step $\downarrow$\\
    \midrule
    Pretrained predictor & 58.15 & 88.94\\
Fixed domain + pretrained & 22.12 & 47.47\\
Random eligible predictor & 46.74 & 81.22\\
\textbf{Error-based selection (ours)} & \textbf{13.03} & \textbf{31.54}\\
Future-error oracle & 11.79 & 27.83\\
 
    \bottomrule
  \end{tabular}}
\end{table}

\vspace{1mm}
\noindent \textbf{Specialization and retention.}
Finally, we examine whether \ours{} can learn a recurring shape without degrading performance on others.
After seven rounds of 50 `I' shape episodes, we evaluate held-out goals across all seven PushObj shapes in Table~\ref{tab:i_only}.
Success rate increases from 47.3\% to 76.7\%, while mean replans decrease from 20.18 to 12.04.
Performance on the other 6 shapes remains nearly unchanged (81.9\% vs.\ 81.8\%).
In contrast, fast-weight inheritance achieves only 38.0\% success on I and 33.8\% on the remaining shapes.
These results demonstrate that \ours{} can improve on a recurring shape while largely preserving performance on others.

\begin{table}[t]
\centering
\caption{\textbf{Shape-specific learning and retention (GD).}
We evaluate success rates (\%) across all seven shapes and mean replans. Each shape has 150 test goals, with failures assigned 30 replans.}
\label{tab:i_only}
\small
\setlength{\tabcolsep}{8pt}
 \adjustbox{max width=\linewidth}{
\begin{tabular}{@{}lccc@{}}
\toprule
Method & I success $\uparrow$ & other success $\uparrow$ & I replans $\downarrow$\\
\midrule
Episodic reset & 47.33 & 81.89 & 20.18\\
Fast-weight inheritance & 38.00 & 33.78 & 22.19\\
\textbf{\ours} & \textbf{76.67} & \textbf{81.78} & \textbf{12.04}\\
 
\bottomrule
\end{tabular}}
\end{table}

\FloatBarrier
\section{Conclusion and Future Work}
\label{sec:conclusion}

\ours turns deployment trajectories into reusable predictive memory by separating fast corrections from slow consolidation. It improves new-goal planning, learns across recurring shape and visual shifts, and transfers to deformable manipulation. The I-only experiment shows that specialization can coexist with retained performance on other shapes, while fast-weight inheritance degrades both. Future work will refine maturity using prediction uncertainty and planning outcomes, and explore compressed replay for longer deployments with bounded storage.

{
  \small
  \bibliographystyle{plainnat}
  \bibliography{references}

\begin{thebibliography}{44}
\providecommand{\natexlab}[1]{#1}
\providecommand{\url}[1]{\texttt{#1}}
\expandafter\ifx\csname urlstyle\endcsname\relax
  \providecommand{\doi}[1]{doi: #1}\else
  \providecommand{\doi}{doi: \begingroup \urlstyle{rm}\Url}\fi

\bibitem[Assran et~al.(2023)Assran, Duval, Misra, Bojanowski, Vincent, Rabbat,
  LeCun, and Ballas]{assran2023ijepa}
Mahmoud Assran, Quentin Duval, Ishan Misra, Piotr Bojanowski, Pascal Vincent,
  Michael Rabbat, Yann LeCun, and Nicolas Ballas.
\newblock Self-supervised learning from images with a joint-embedding
  predictive architecture.
\newblock In \emph{IEEE/CVF Conference on Computer Vision and Pattern
  Recognition}, 2023.

\bibitem[Assran et~al.(2025)Assran, Bardes, Fan, Garrido,
  et~al.]{assran2025vjepa2}
Mido Assran, Adrien Bardes, David Fan, Quentin Garrido, et~al.
\newblock {V-JEPA 2}: Self-supervised video models enable understanding,
  prediction and planning.
\newblock \emph{arXiv preprint arXiv:2506.09985}, 2025.

\bibitem[Bardes et~al.(2022)Bardes, Ponce, and LeCun]{bardes2022vicreg}
Adrien Bardes, Jean Ponce, and Yann LeCun.
\newblock {VICReg}: Variance-invariance-covariance regularization for
  self-supervised learning.
\newblock In \emph{International Conference on Learning Representations}, 2022.

\bibitem[Bardes et~al.(2024)Bardes, Garrido, Ponce, Chen, Rabbat, LeCun,
  Assran, and Ballas]{bardes2024vjepa}
Adrien Bardes, Quentin Garrido, Jean Ponce, Xinlei Chen, Michael Rabbat, Yann
  LeCun, Mahmoud Assran, and Nicolas Ballas.
\newblock Revisiting feature prediction for learning visual representations
  from video.
\newblock \emph{arXiv preprint arXiv:2404.08471}, 2024.

\bibitem[Caron et~al.(2021)Caron, Touvron, Misra, J{\'e}gou, Mairal,
  Bojanowski, and Joulin]{caron2021dino}
Mathilde Caron, Hugo Touvron, Ishan Misra, Herv{\'e} J{\'e}gou, Julien Mairal,
  Piotr Bojanowski, and Armand Joulin.
\newblock Emerging properties in self-supervised vision transformers.
\newblock In \emph{IEEE/CVF International Conference on Computer Vision}, 2021.

\bibitem[Chua et~al.(2018)Chua, Calandra, McAllister, and Levine]{chua2018pets}
Kurtland Chua, Roberto Calandra, Rowan McAllister, and Sergey Levine.
\newblock Deep reinforcement learning in a handful of trials using
  probabilistic dynamics models.
\newblock In \emph{Advances in Neural Information Processing Systems}, 2018.

\bibitem[Fang et~al.(2023)Fang, Xu, Wang, Yang, Wang, and Zhou]{fang2023pvd}
Shuangkang Fang, Weixin Xu, Heng Wang, Yi~Yang, Yufeng Wang, and Shuchang Zhou.
\newblock One is all: Bridging the gap between neural radiance fields
  architectures with progressive volume distillation.
\newblock In \emph{Proceedings of the AAAI Conference on Artificial
  Intelligence}, volume~37, pages 597--605, 2023.

\bibitem[Fang et~al.(2026)Fang, Wang, Yang, Xu, Wang, Ding, and
  Zhou]{fang2026pvdal}
Shuangkang Fang, Yufeng Wang, Yi~Yang, Weixin Xu, Heng Wang, Wenrui Ding, and
  Shuchang Zhou.
\newblock Progressive volume distillation with active learning for efficient
  nerf architecture conversion.
\newblock \emph{International Journal of Computer Vision}, 134\penalty0
  (5):\penalty0 228, 2026.

\bibitem[Grill et~al.(2020)Grill, Strub, Altch{\'e}, Tallec, Richemond,
  Buchatskaya, Doersch, Avila~Pires, Guo, Gheshlaghi~Azar, Piot, Kavukcuoglu,
  Munos, and Valko]{grill2020byol}
Jean-Bastien Grill, Florian Strub, Florent Altch{\'e}, Corentin Tallec,
  Pierre~H. Richemond, Elena Buchatskaya, Carl Doersch, Bernardo Avila~Pires,
  Zhaohan~Daniel Guo, Mohammad Gheshlaghi~Azar, Bilal Piot, Koray Kavukcuoglu,
  R{\'e}mi Munos, and Michal Valko.
\newblock Bootstrap your own latent: A new approach to self-supervised
  learning.
\newblock In \emph{Advances in Neural Information Processing Systems}, 2020.

\bibitem[Ha and Schmidhuber(2018)]{ha2018worldmodels}
David Ha and J{\"u}rgen Schmidhuber.
\newblock World models.
\newblock \emph{arXiv preprint arXiv:1803.10122}, 2018.

\bibitem[Hafner et~al.(2019)Hafner, Lillicrap, Fischer, Villegas, Ha, Lee, and
  Davidson]{hafner2019planet}
Danijar Hafner, Timothy Lillicrap, Ian Fischer, Ruben Villegas, David Ha,
  Honglak Lee, and James Davidson.
\newblock Learning latent dynamics for planning from pixels.
\newblock In \emph{International Conference on Machine Learning}, pages
  2555--2565, 2019.

\bibitem[Hafner et~al.(2020)Hafner, Lillicrap, Ba, and
  Norouzi]{hafner2020dreamer}
Danijar Hafner, Timothy Lillicrap, Jimmy Ba, and Mohammad Norouzi.
\newblock Dream to control: Learning behaviors by latent imagination.
\newblock In \emph{International Conference on Learning Representations}, 2020.

\bibitem[Hafner et~al.(2021)Hafner, Lillicrap, Norouzi, and
  Ba]{hafner2021dreamerv2}
Danijar Hafner, Timothy Lillicrap, Mohammad Norouzi, and Jimmy Ba.
\newblock Mastering atari with discrete world models.
\newblock In \emph{International Conference on Learning Representations}, 2021.

\bibitem[Hafner et~al.(2025)Hafner, Pasukonis, Ba, and
  Lillicrap]{hafner2025dreamerv3}
Danijar Hafner, Jurgis Pasukonis, Jimmy Ba, and Timothy Lillicrap.
\newblock Mastering diverse control tasks through world models.
\newblock \emph{Nature}, 2025.

\bibitem[Hansen et~al.(2024)Hansen, Su, and Wang]{hansen2024tdmpc2}
Nicklas Hansen, Hao Su, and Xiaolong Wang.
\newblock {TD-MPC2}: Scalable, robust world models for continuous control.
\newblock In \emph{International Conference on Learning Representations}, 2024.

\bibitem[Hansen et~al.(2022)Hansen, Su, and Wang]{hansen2022tdmpc}
Nicklas~A. Hansen, Hao Su, and Xiaolong Wang.
\newblock Temporal difference learning for model predictive control.
\newblock In \emph{International Conference on Machine Learning}, pages
  8387--8406, 2022.

\bibitem[He et~al.(2016)He, Zhang, Ren, and Sun]{he2016resnet}
Kaiming He, Xiangyu Zhang, Shaoqing Ren, and Jian Sun.
\newblock Deep residual learning for image recognition.
\newblock In \emph{IEEE Conference on Computer Vision and Pattern Recognition},
  pages 770--778, 2016.

\bibitem[Kaiser et~al.(2020)Kaiser, Babaeizadeh, Milos, Osinski, Campbell,
  Czechowski, Erhan, Finn, Kozakowski, Levine, Mohiuddin, Sepassi, Tucker, and
  Michalewski]{kaiser2020simple}
Lukasz Kaiser, Mohammad Babaeizadeh, Piotr Milos, Blazej Osinski, Roy~H.
  Campbell, Konrad Czechowski, Dumitru Erhan, Chelsea Finn, Piotr Kozakowski,
  Sergey Levine, Afroz Mohiuddin, Ryan Sepassi, George Tucker, and Henryk
  Michalewski.
\newblock Model-based reinforcement learning for atari.
\newblock In \emph{International Conference on Learning Representations}, 2020.

\bibitem[Kirkpatrick et~al.(2017)Kirkpatrick, Pascanu, Rabinowitz, Veness,
  Desjardins, Rusu, Milan, Quan, Ramalho, Grabska-Barwinska, Hassabis, Clopath,
  Kumaran, and Hadsell]{kirkpatrick2017ewc}
James Kirkpatrick, Razvan Pascanu, Neil Rabinowitz, Joel Veness, Guillaume
  Desjardins, Andrei~A. Rusu, Kieran Milan, John Quan, Tiago Ramalho, Agnieszka
  Grabska-Barwinska, Demis Hassabis, Claudia Clopath, Dharshan Kumaran, and
  Raia Hadsell.
\newblock Overcoming catastrophic forgetting in neural networks.
\newblock \emph{Proceedings of the National Academy of Sciences}, 114\penalty0
  (13):\penalty0 3521--3526, 2017.

\bibitem[LeCun(2022)]{lecun2022path}
Yann LeCun.
\newblock A path towards autonomous machine intelligence.
\newblock OpenReview position paper, 2022.

\bibitem[Li and Hoiem(2016)]{li2016lwf}
Zhizhong Li and Derek Hoiem.
\newblock Learning without forgetting.
\newblock In \emph{European Conference on Computer Vision}, pages 614--629,
  2016.

\bibitem[Lin et~al.(2021)Lin, Wang, Olkin, and Held]{lin2020softgym}
Xingyu Lin, Yufei Wang, Jake Olkin, and David Held.
\newblock {SoftGym}: Benchmarking deep reinforcement learning for deformable
  object manipulation.
\newblock In \emph{Proceedings of the 2020 Conference on Robot Learning}, pages
  432--448, 2021.

\bibitem[Lopez-Paz and Ranzato(2017)]{lopezpaz2017gem}
David Lopez-Paz and Marc'Aurelio Ranzato.
\newblock Gradient episodic memory for continual learning.
\newblock In \emph{Advances in Neural Information Processing Systems}, 2017.

\bibitem[Maes et~al.(2026)Maes, Le~Lidec, Scieur, LeCun, and
  Balestriero]{maes2026lewm}
Lucas Maes, Quentin Le~Lidec, Damien Scieur, Yann LeCun, and Randall
  Balestriero.
\newblock {LeWorldModel}: Stable end-to-end joint-embedding predictive
  architecture from pixels.
\newblock \emph{arXiv preprint arXiv:2603.19312}, 2026.

\bibitem[McClelland et~al.(1995)McClelland, McNaughton, and
  O'Reilly]{mcclelland1995cls}
James~L. McClelland, Bruce~L. McNaughton, and Randall~C. O'Reilly.
\newblock Why there are complementary learning systems in the hippocampus and
  neocortex: Insights from the successes and failures of connectionist models
  of learning and memory.
\newblock \emph{Psychological Review}, 102\penalty0 (3):\penalty0 419--457,
  1995.

\bibitem[Micheli et~al.(2023)Micheli, Alonso, and Fleuret]{micheli2023iris}
Vincent Micheli, Eloi Alonso, and Fran{\c{c}}ois Fleuret.
\newblock Transformers are sample-efficient world models.
\newblock In \emph{International Conference on Learning Representations}, 2023.

\bibitem[Nair et~al.(2023)Nair, Rajeswaran, Kumar, Finn, and
  Gupta]{nair2022r3m}
Suraj Nair, Aravind Rajeswaran, Vikash Kumar, Chelsea Finn, and Abhinav Gupta.
\newblock {R3M}: A universal visual representation for robot manipulation.
\newblock In \emph{Proceedings of the 6th Conference on Robot Learning}, pages
  892--909, 2023.

\bibitem[Oquab et~al.(2024)Oquab, Darcet, Moutakanni, Vo, Szafraniec, Khalidov,
  Fernandez, Haziza, Massa, El-Nouby, Assran, Ballas, Galuba, Howes, Huang, Li,
  Misra, Rabbat, Sharma, Synnaeve, Xu, J{\'e}gou, Mairal, Labatut, Joulin, and
  Bojanowski]{oquab2024dinov2}
Maxime Oquab, Timoth{\'e}e Darcet, Th{\'e}o Moutakanni, Huy Vo, Marc
  Szafraniec, Vasil Khalidov, Pierre Fernandez, Daniel Haziza, Francisco Massa,
  Alaaeldin El-Nouby, Mahmoud Assran, Nicolas Ballas, Wojciech Galuba, Russell
  Howes, Po-Yao Huang, Shang-Wen Li, Ishan Misra, Michael Rabbat, Vasu Sharma,
  Gabriel Synnaeve, Hu~Xu, Herv{\'e} J{\'e}gou, Julien Mairal, Patrick Labatut,
  Armand Joulin, and Piotr Bojanowski.
\newblock {DINOv2}: Learning robust visual features without supervision.
\newblock \emph{Transactions on Machine Learning Research}, 2024.

\bibitem[Rolnick et~al.(2019)Rolnick, Ahuja, Schwarz, Lillicrap, and
  Wayne]{rolnick2019replay}
David Rolnick, Arun Ahuja, Jonathan Schwarz, Timothy Lillicrap, and Gregory
  Wayne.
\newblock Experience replay for continual learning.
\newblock In \emph{Advances in Neural Information Processing Systems}, 2019.

\bibitem[Rusu et~al.(2016)Rusu, Rabinowitz, Desjardins, Soyer, Kirkpatrick,
  Kavukcuoglu, Pascanu, and Hadsell]{rusu2016progressive}
Andrei~A. Rusu, Neil~C. Rabinowitz, Guillaume Desjardins, Hubert Soyer, James
  Kirkpatrick, Koray Kavukcuoglu, Razvan Pascanu, and Raia Hadsell.
\newblock Progressive neural networks.
\newblock \emph{arXiv preprint arXiv:1606.04671}, 2016.

\bibitem[Schrittwieser et~al.(2020)Schrittwieser, Antonoglou, Hubert, Simonyan,
  Sifre, Schmitt, Guez, Lockhart, Hassabis, Graepel, Lillicrap, and
  Silver]{schrittwieser2020muzero}
Julian Schrittwieser, Ioannis Antonoglou, Thomas Hubert, Karen Simonyan,
  Laurent Sifre, Simon Schmitt, Arthur Guez, Edward Lockhart, Demis Hassabis,
  Thore Graepel, Timothy Lillicrap, and David Silver.
\newblock Mastering atari, go, chess and shogi by planning with a learned
  model.
\newblock \emph{Nature}, 588:\penalty0 604--609, 2020.

\bibitem[Schwarzer et~al.(2021)Schwarzer, Anand, Goel, Hjelm, Courville, and
  Bachman]{schwarzer2021spr}
Max Schwarzer, Ankesh Anand, Rishab Goel, R.~Devon Hjelm, Aaron Courville, and
  Philip Bachman.
\newblock Data-efficient reinforcement learning with self-predictive
  representations.
\newblock In \emph{International Conference on Learning Representations}, 2021.

\bibitem[Sun et~al.(2020)Sun, Wang, Liu, Miller, Efros, and Hardt]{sun2020ttt}
Yu~Sun, Xiaolong Wang, Zhuang Liu, John Miller, Alexei~A. Efros, and Moritz
  Hardt.
\newblock Test-time training with self-supervision for generalization under
  distribution shifts.
\newblock In \emph{International Conference on Machine Learning}, 2020.

\bibitem[van~den Oord et~al.(2018)van~den Oord, Li, and Vinyals]{oord2018cpc}
Aaron van~den Oord, Yazhe Li, and Oriol Vinyals.
\newblock Representation learning with contrastive predictive coding.
\newblock \emph{arXiv preprint arXiv:1807.03748}, 2018.

\bibitem[Wang et~al.(2021)Wang, Shelhamer, Liu, Olshausen, and
  Darrell]{wang2021tent}
Dequan Wang, Evan Shelhamer, Shaoteng Liu, Bruno Olshausen, and Trevor Darrell.
\newblock Tent: Fully test-time adaptation by entropy minimization.
\newblock In \emph{International Conference on Learning Representations}, 2021.

\bibitem[Wang et~al.(2022)Wang, Fink, Van~Gool, and Dai]{wang2022cotta}
Qin Wang, Olga Fink, Luc Van~Gool, and Dengxin Dai.
\newblock Continual test-time domain adaptation.
\newblock In \emph{IEEE/CVF Conference on Computer Vision and Pattern
  Recognition}, 2022.

\bibitem[Wang et~al.(2026{\natexlab{a}})Wang, Bounou, LeCun, and
  Ren]{wang2026adajepa}
Ying Wang, Oumayma Bounou, Yann LeCun, and Mengye Ren.
\newblock {AdaJEPA}: An adaptive latent world model.
\newblock \emph{arXiv preprint arXiv:2606.32026}, 2026{\natexlab{a}}.

\bibitem[Wang et~al.(2026{\natexlab{b}})Wang, Bounou, Zhou, Balestriero,
  Rudner, LeCun, and Ren]{wang2026straightening}
Ying Wang, Oumayma Bounou, Gaoyue Zhou, Randall Balestriero, Tim G.~J. Rudner,
  Yann LeCun, and Mengye Ren.
\newblock Temporal straightening for latent planning.
\newblock In \emph{International Conference on Machine Learning},
  2026{\natexlab{b}}.

\bibitem[Watter et~al.(2015)Watter, Springenberg, Boedecker, and
  Riedmiller]{watter2015e2c}
Manuel Watter, Jost~Tobias Springenberg, Joschka Boedecker, and Martin
  Riedmiller.
\newblock Embed to control: A locally linear latent dynamics model for control
  from raw images.
\newblock In \emph{Advances in Neural Information Processing Systems}, 2015.

\bibitem[Ye et~al.(2021)Ye, Liu, Kurutach, Abbeel, and
  Gao]{ye2021efficientzero}
Weirui Ye, Shaohuai Liu, Thanard Kurutach, Pieter Abbeel, and Yang Gao.
\newblock Mastering atari games with limited data.
\newblock In \emph{Advances in Neural Information Processing Systems}, 2021.

\bibitem[Yin et~al.(2026)Yin, Yin, Chen, Liu, Li, and Lin]{yin2026ddpwm}
Shicheng Yin, Kaixuan Yin, Weixing Chen, Yang Liu, Guanbin Li, and Liang Lin.
\newblock {DDP-WM}: Disentangled dynamics prediction for efficient world
  models.
\newblock \emph{arXiv preprint arXiv:2602.01780}, 2026.

\bibitem[Zbontar et~al.(2021)Zbontar, Jing, Misra, LeCun, and
  Deny]{zbontar2021barlow}
Jure Zbontar, Li~Jing, Ishan Misra, Yann LeCun, and Stephane Deny.
\newblock Barlow twins: Self-supervised learning via redundancy reduction.
\newblock In \emph{International Conference on Machine Learning}, pages
  12310--12320, 2021.

\bibitem[Zhang et~al.(2026)Zhang, Ye, Agarwal, Mahjoub, Moradi~Pari, Khashabi,
  Shu, and Tadiparthi]{zhang2026jepattt}
Zheyuan Zhang, Suyu Ye, Nakul Agarwal, Hossein~Nourkhiz Mahjoub, Ehsan
  Moradi~Pari, Daniel Khashabi, Tianmin Shu, and Vaishnav Tadiparthi.
\newblock {JEPA-TTT}: Persistent test-time training of latent world models for
  planning under dynamics shifts.
\newblock \emph{arXiv preprint arXiv:2610.00722}, 2026.

\bibitem[Zhou et~al.(2025)Zhou, Pan, LeCun, and Pinto]{zhou2024dinowm}
Gaoyue Zhou, Hengkai Pan, Yann LeCun, and Lerrel Pinto.
\newblock {DINO-WM}: World models on pre-trained visual features enable
  zero-shot planning.
\newblock In \emph{International Conference on Machine Learning}, pages
  79115--79135, 2025.

\end{thebibliography}
}

\clearpage
\appendix
\setcounter{figure}{0}
\setcounter{table}{0}
\setcounter{equation}{0}
\renewcommand{\thefigure}{\Roman{figure}}
\renewcommand{\thetable}{\Roman{table}}
\renewcommand{\theequation}{\Roman{equation}}
\renewcommand{\thealgorithm}{\Roman{algorithm}}
\renewcommand{\theHfigure}{app.\arabic{figure}}
\renewcommand{\theHtable}{app.\arabic{table}}
\renewcommand{\theHequation}{app.\arabic{equation}}
\renewcommand{\theHalgorithm}{app.\arabic{algorithm}}

\section*{Supplementary Material}
\noindent In this supplementary material, we provide additional implementation details, evaluation protocols, and analyses of SE-JEPA.
Figures, tables, and equations in this supplement are numbered with Roman numerals (\eg, Table~I), whereas Arabic numerals (\eg, Table~1) refer to the main text.
The supplement is organized as follows:
\begin{itemize}[label=\textbullet,leftmargin=1em,itemsep=0pt,topsep=3pt,parsep=0pt]
  \item \textbf{Appendix~\ref{app:implementation}} details fast adaptation, slow consolidation, label-free domain discovery and merging, memory management, and deployment costs.
  \item \textbf{Appendix~\ref{app:protocols}} describes the evaluation protocols for PushObj, PushT, and deformable manipulation, including dataset splits, held-out testing, planning budgets, and evaluation metrics.
  \item \textbf{Appendix~\ref{app:baselines}} documents model initialization, baseline implementations, evaluation settings, and sources of reported results.
  \item \textbf{Appendix~\ref{app:additional}} provides per-condition results, learning across recurring shifts, deformable manipulation analyses, memory-design ablations, predictor retrieval, and shape-specific continual learning.
  \item \textbf{Appendix~\ref{app:qualitative}} presents additional planning trajectories and qualitative comparisons on PushObj, PushT, and deformable tasks, including execution sequences and prediction-error analyses.
\end{itemize}

\section{Implementation Details}
\label{app:implementation}

\subsection{Fast Adaptation and Replay Consolidation}
On PushObj and PushT, we follow AdaJEPA~\maincite{wang2026adajepa}{1} and adapt episode-local copies of the encoder and predictor using recently observed transitions.
At each replan, one Adam step updates the predictor's last block and the encoder head.
These working copies reset between episodes, while the pretrained reference $(E_0, P_0)$ remains fixed.
Slow consolidation trains a separate full predictor for each predictive domain, initialized from $P_0$.
Episode trajectories are re-encoded with $E_0$ and stored as latent sequences.
Each replay step samples up to eight episodes with replacement and a random window of up to six frames from each.
The objective averages one-step prediction errors on observation features, with actions as conditioning inputs.
A standard write triggers 15 Adam steps; initializing a predictor from $n$ stored episodes uses $15n$ steps, and merging adds updates for the absorbed experience.
Table~\ref{tab:hyperparams} summarizes the configuration.

\begin{table}[t]
  \centering
  \caption{\textbf{Hyperparameter settings for PushObj and PushT datasets.}
  The seven-round shape stream in \appendixref{app:additional} uses a slow learning rate of $2\times10^{-4}$.}
  \label{tab:hyperparams}
  \small
  \begin{tabular}{lr}
    \toprule
    Setting & Value\\
    \midrule
    Fast predictor learning rate & $5\times10^{-4}$\\
    Fast encoder-head learning rate & $10^{-5}$\\
    Fast steps per replan & 1\\
    Recent adaptation buffer & 5 transitions\\
    Slow learning rate & $10^{-4}$\\
    Slow steps per standard write & 15\\
    Slow batch size (episodes) & 8\\
    Slow sampled window (frames) & up to 6\\
    Familiarity threshold $\tau_f$ & 0.0146\\
    Existing-domain cosine threshold & 0.10\\
    New-domain cosine threshold & 0.05\\
    Episodes to confirm a domain & 3\\
    Pending-pool capacity & 40\\
    Minimum episodes to train a learner & 3\\
    Retrieval window & first 3 replans\\
    Maturity history length $m$ & 3\\
    Maturity error-ratio threshold $\tau_m$ & 0.9\\
    Write rejection ratio & 1.2\\
    Merge evidence per direction & 2 episodes\\
    Merge relative tolerance & 0.5\\
    Fragment absorption interval & 25 episodes\\
    Absorption error-ratio threshold & 0.8\\
    Fragment-size limit & $<6$ episodes\\
    \bottomrule
  \end{tabular}
\end{table}

\subsection{Deployment Procedure}
Algorithm~\ref{alg:sejepa} summarizes a deployment episode.
During the first three replans, observed transitions guide familiarity testing and predictor retrieval.
Familiarity is measured by the geometric mean of the reference predictor's newest-frame errors over up to the first three transitions; its threshold is calibrated on the PushObj backbone and reused unchanged for PushT.
Retrieval replaces the predictor base while preserving the accumulated fast offset.
Validation always precedes replay on the current trajectory.
During held-out evaluation on PushObj and PushT, all persistent memory state, including validation histories, remains fixed; retrieval and fast adaptation remain active.

\begin{algorithm}[t]
\caption{A deployment-learning episode of \ours{}}
\label{alg:sejepa}
\begin{algorithmic}[1]
  \Require Frozen $(E_0,P_0)$, memory $\memory$, initial observation $o$, goal $g$
  \State Reset episode-local parameters to $(E_0,P_0)$
  \State Initialize recent buffer $\mathcal{B}$ and episode trajectory $\tau$
  \While{budget remains and the goal has not been reached}
    \State Plan an action sequence with the current model
    \State Execute its first block and observe the next state
    \State Append observed transitions to $\mathcal{B}$ and $\tau$
    \If{within the first three replans}
      \State Form candidates from $P_0$ and mature slow predictors
      \State Restrict candidates to $P_0$ if the episode is familiar
      \State Select the predictor with the lowest observed prediction error
      \State Replace the predictor base, retaining the fast offset
    \EndIf
    \State Adapt the working model on recent transitions
  \EndWhile
  \State Encode $\tau$ with frozen $E_0$
  \State Measure existing predictors' rollout errors on $\tau$ before replay
  \If{the episode is unfamiliar}
    \State Propose a domain using retrieval or an update fingerprint
    \State Reject incompatible writes or retain pending evidence
    \If{a domain is assigned or newly confirmed}
      \State Record validation if a learner already exists
      \State Append latent experience and update the slow predictor
      \State Check domain merging and periodic absorption
    \EndIf
  \EndIf
  \State Discard episode-local parameters
\end{algorithmic}
\end{algorithm}

\subsection{Domain Assignment and Discovery}
An update fingerprint is the normalized concatenation of predictor and encoder-head parameter changes.
Domain prototypes accumulate fingerprints only from episodes planned from the pretrained reference.
For trajectory assignment, the retrieved predictor's domain is considered first, subject to the pre-replay error-ratio check.
Otherwise, candidate domains are tested in descending order of cosine similarity.

Unassigned episodes enter a bounded pending pool.
A new domain is confirmed when the incoming fingerprint has cosine similarity above the new-domain threshold with at least two pending fingerprints; pairwise agreement among the pending fingerprints is not required.
The matched episodes and the incoming episode jointly initialize its replay buffer.
Both assignment and domain discovery operate without shape or appearance labels.

\subsection{Domain Merging and Absorption}
Merging uses prediction errors measured before replay on unfamiliar trajectories.
Let $G_{A\to B}$ denote the geometric mean error ratio of predictor $B$ on recent episodes assigned to domain $A$, and $G_{A\to A}$ the corresponding ratio for predictor $A$.
Domains merge when
\begin{equation}
  G_{A\to B}\leq \min\{1,(1+\epsilon_m)G_{A\to A}\}
\end{equation}
and the symmetric condition holds for $B\to A$, with at least two jointly evaluated episodes in each direction.
The larger domain retains its predictor and combines both replay buffers.

Every 25 episodes, we also test whether small immature domains can be absorbed into mature domains.
Eligible domains contain fewer than six episodes and fewer than three maturity measurements; the receiving predictor must achieve an error ratio of at most 0.8 on their stored experience.
Pending-pool absorption and the additional absolute-error write gate are disabled for PushObj and PushT.

\subsection{Runtime and Memory Analysis}
\label{app:deployment_cost}

\vspace{1mm} \noindent \textbf{Profiling protocol.}
The training process for \ours{} is conducted on a server equipped with eight NVIDIA A100 GPUs.
For profiling, we implement Frozen, AdaJEPA, and \ours{} on a single NVIDIA RTX 4090 using both GD and CEM.
All methods share the same planner implementation and evaluate ten matched PushT tasks under seven visual conditions.
Three timing repetitions per method, planner, and task--condition pair yield 1,260 executions in total.
For \ours{}, we reuse memory learned over five rounds, with retrieval and fast adaptation active and persistent memory fixed.
Timing includes planning, adaptation, retrieval, simulator execution, and evaluation, but excludes initialization and two warm-up episodes per run.
We synchronize CUDA operations, capture repeated GD optimization with CUDA graphs during warm-up, and average timings within each task--condition pair.

\vspace{1mm} \noindent \textbf{Runtime and planning performance.}
Relative to AdaJEPA, \ours{} improves success by 34.3 percentage points under GD and 35.7 points under CEM, while reducing mean episode time by 26.7\% and 27.6\%, respectively (\paperxref{Table}{tab:deployment_cost}{}).
The runtime reduction primarily reflects fewer replans: planning costs remain similar across methods at approximately 0.30 seconds per replan under GD and 0.44--0.45 seconds under CEM.
The breakdown in \paperxref{Table}{tab:deployment_phases}{} shows mean retrieval costs of 0.094 and 0.118 seconds per episode, respectively.
Peak allocated GPU memory for \ours{} is 999\,MB under GD and 3{,}361\,MB under CEM.

\begin{table*}[t]
  \centering
  \caption{\textbf{Planning performance and deployment cost on a single RTX 4090.}
  We report success (\%), mean MPC replans and episode time (s), and peak allocated and reserved GPU memory (MB) on matched PushT tasks.}
  \label{tab:deployment_cost}
  \small
  \begin{tabular}{@{}llccccc@{}}
    \toprule
    Planner & Method & Success $\uparrow$ & Replans $\downarrow$ & Mean time $\downarrow$ & Allocated & Reserved\\
    \midrule
    GD & Frozen & 21.43 & 24.94 & 17.88 & \textbf{225} & \textbf{322}\\
GD & AdaJEPA & 32.86 & 23.56 & 18.30 & 322 & 387\\
GD & \textbf{\ours} & \textbf{67.14} & \textbf{15.91} & \textbf{13.42} & 999 & 1,095\\
\midrule
CEM & Frozen & 22.86 & 24.71 & 21.26 & \textbf{2,856} & \textbf{3,221}\\
CEM & AdaJEPA & 35.71 & 23.09 & 21.35 & 2,888 & 3,232\\
CEM & \textbf{\ours} & \textbf{71.43} & \textbf{15.69} & \textbf{15.45} & 3,361 & 4,198\\
 
    \bottomrule
  \end{tabular}
\end{table*}

\begin{table*}[t]
  \centering
  \caption{\textbf{Episode runtime breakdown (s).}
  Adaptation includes prediction diagnostics, and simulator time includes evaluation.
  Total runtimes in \paperxref{Table}{tab:deployment_cost}{} also include reset and bookkeeping.}
  \label{tab:deployment_phases}
  \small
  \begin{tabular}{@{}llcccc@{}}
    \toprule
    Planner & Method & Planning & Adaptation & Retrieval & Simulator/evaluation\\
    \midrule
    GD & Frozen & 7.568 & \textbf{0.000} & \textbf{0.000} & 10.267\\
GD & AdaJEPA & 7.140 & 1.722 & \textbf{0.000} & 9.401\\
GD & \textbf{\ours} & \textbf{4.782} & 3.188 & 0.094 & \textbf{5.329}\\
\midrule
CEM & Frozen & 11.007 & \textbf{0.000} & \textbf{0.000} & 10.214\\
CEM & AdaJEPA & 10.355 & 1.730 & \textbf{0.000} & 9.226\\
CEM & \textbf{\ours} & \textbf{6.905} & 3.237 & 0.118 & \textbf{5.164}\\
 
    \bottomrule
  \end{tabular}
\end{table*}

\vspace{1mm} \noindent \textbf{Consolidation and storage.}
For $K$ predictive domains, persistent memory stores $K$ slow predictors alongside the frozen reference, with replay storage proportional to the retained latent transitions.
\paperxref{Table}{tab:deployment_memory}{} summarizes six five-round learning streams of 1,750 episodes each.
Final memories retain 733--766 trajectories, and cumulative slow optimization ranges from 11,280 to 15,180 steps per stream.
These counts are reconstructed from recorded update events and include predictor initialization, standard writes, merging, and absorption.
The table reports replay storage and full saved-state sizes separately; the latter also includes predictor and optimizer tensors, fingerprints, and pending experience.
These updates belong to the preceding learning stage, while the runtime measurements above cover subsequent memory reuse.

\begin{table*}[t]
  \centering
  \caption{\textbf{Persistent memory after five deployment rounds.}
  For each stream, we report active/mature predictor counts, retained trajectories, replay and full saved-state sizes (MB), and cumulative slow optimization steps.}
  \label{tab:deployment_memory}
  \small
  \begin{tabular}{@{}lcccccc@{}}
    \toprule
    Planner & Stream & Active/mature & Trajectories & Replay MB & State MB & Slow steps\\
    \midrule
    GD & 1 & 4 / 1 & 766 & 24.0 & 550.9 & 13,455\\
GD & 2 & 2 / 1 & 747 & 23.6 & 346.4 & 13,155\\
GD & 3 & 5 / 2 & 742 & 21.5 & 604.0 & 15,180\\
CEM & 1 & 3 / 3 & 733 & 22.5 & 476.9 & 11,280\\
CEM & 2 & 5 / 3 & 750 & 22.9 & 606.0 & 13,080\\
CEM & 3 & 4 / 2 & 740 & 21.8 & 548.0 & 12,960\\
 
    \bottomrule
  \end{tabular}
\end{table*}

\section{Evaluation Protocols}
\label{app:protocols}

\subsection{Shape Shifts}
PushObj covers seven shapes: T, L, Z, and plus are seen during pretraining, whereas I, square, and small tee are unseen.
Each learning stream runs five rounds with 50 episodes per shape and round.
Held-out testing uses target seeds 2000, 2100, and 2200, each with 350 episodes, and disables memory writes.
Single-seed baseline runs use seed 2000.

For the paired comparison in \paperxref{Table}{tab:planners}{}, each planner has three learned memory states, and every state is evaluated on all three test seeds.
We average over episodes and states within each seed and report the standard deviation across seed-level means.
The \ours{} PushObj entries in \paperxref{Table}{tab:main}{} instead report fifth-round online performance with memory writes active: 844 of 1,050 episodes overall and 329 of 450 on unseen shapes.

\subsection{Visual Shifts}
PushT covers the original T condition and six shifts: blur, salt-and-pepper noise, darkness, red agent, red object, and red goal.
Three five-round learning chains resample 350 targets in every round.
Held-out testing uses three target seeds with 50 episodes per condition.
Source-episode identifiers confirm that held-out sources never appear in the learning streams and that the three test seeds are disjoint.

The color curve in \paperxref{Fig.}{fig:deployment}{}(b) pools the three recoloring conditions, which share the same 50 source targets within a chain and round.
For each chain and round, we divide the pooled successes by 150 episodes and plot the mean over three chains.
The VIS50 baseline comparison evaluates the same 50 initial/goal pairs under each of the six shifts with planning seed 100.

\subsection{Success and Interaction Counts}
On PushObj and PushT, GD uses 100 optimization steps, and CEM uses 10 iterations with 200 samples and 30 elites.
Success requires a joint pusher--object position error below 20 simulator units and an object orientation error below $\pi/9$.
Each MPC replan executes five physical actions, and an episode allows at most $T_{\max}=30$ replans.
For episode $i$ with success indicator $S_i$ and first successful replan $T_i$, the mean replan count is
\begin{equation}
  \overline{T} = \frac{1}{N}\sum_{i=1}^{N}
  \big[\, S_i T_i + (1-S_i)\, T_{\max} \big]
  = T_{\max} - \sum_{t=1}^{T_{\max}-1} F(t),
\end{equation}
where $F(t)$ is the fraction of episodes solved by replan $t$.
\paperxref{Fig.}{fig:efficiency_main}{} plots $F(t)$; \ours{} pulls ahead of AdaJEPA within the first four replans and keeps the lead.
DINO-WM executes 25 actions per replan on PushT CEM, so \paperxref{Table}{tab:main}{} converts its mean to five-action units: $4.41\times25/5=22.05$.

\subsection{Deformable Manipulation}
\paperxref{Table}{tab:deform_hyperparams}{} lists the rope and granular settings.
In single-material consolidation, every pass revisits the same targets in the same order with the same planning seeds.
For the unseen-rope test, we freeze the predictor saved after the first pass and compare it with the frozen reference on 91 instances absent from the learning stream.
The one-pass predictor lowers CD on 56 of the 91 instances.
A paired percentile bootstrap over instances (20,000 resamples) gives a 95\% interval of [0.162, 0.470] for the mean CD reduction of 0.314.

\begin{table}[t]
\centering
\caption{\textbf{Rope and Granular hyperparameters.} Planning settings apply to both materials. Single-material consolidation uses a separate slow predictor for each material; mixed-material learning retains one memory across material switches.}
\label{tab:deform_hyperparams}
\footnotesize
\setlength{\tabcolsep}{3pt}
\begin{tabular}{@{}p{0.59\columnwidth}p{0.35\columnwidth}@{}}
\toprule
Setting & Value\\
\midrule
\multicolumn{2}{@{}l}{\textbf{Planning and evaluation}}\\
Backbone & DINO-WM\\
Planner & CEM\\
CEM samples / elites / iterations & 300 / 30 / 10\\
MPC replans / pushes per replan & 3 / 5\\
Maximum pushes per episode & 15\\
Evaluation metric & Final CD $\downarrow$\\
\midrule
\multicolumn{2}{@{}l}{\textbf{Single-material consolidation}}\\
Instances per material / passes & 50 / 4\\
Slow predictors per material & 1\\
Fast adaptation & Disabled\\
Slow learning rate & $10^{-4}$\\
Alternative slow learning rate & $2\times10^{-4}$\\
Unseen Rope instances & 91\\
Frozen predictor for unseen test & After pass 1\\
\midrule
\multicolumn{2}{@{}l}{\textbf{Mixed-material learning}}\\
Episodes per material / round & 50\\
Rounds / total episodes & 3 / 300\\
Consecutive episodes per material & 10\\
Optimizer & Adam\\
Fast steps per replan & 1\\
Fast predictor / encoder-head LR & $5\times10^{-4}$ / $10^{-5}$\\
Recent adaptation buffer & 5 transitions\\
Slow learning rate & $2\times10^{-4}$\\
Slow steps per standard write & 15\\
Slow batch size (episodes) & 8\\
Familiarity threshold & 0.117\\
Maturity history / error ratio & 3 / 0.9\\
Absolute write-error ratio limit & 20\\
\bottomrule
\end{tabular}
\end{table}

\begin{table*}[t]
 \centering
 \caption{\textbf{Held-out results with both planners.} Success (\%) and mean MPC replans (Steps) on the paired test sets after five learning rounds. PushObj All averages seven shapes and Unseen averages I, square, and small tee. PushT All averages seven conditions, Shift excludes Original, and Steps averages all seven conditions.}
 \label{tab:planners}
 \small
 \begin{tabular}{llcccccc}
 \toprule
 & & \multicolumn{3}{c}{PushObj} & \multicolumn{3}{c}{PushT}\\
 \cmidrule(lr){3-5}\cmidrule(lr){6-8}
 Planner & Method & All $\uparrow$ & Unseen $\uparrow$ & Steps $\downarrow$ & All $\uparrow$ & Shift $\uparrow$ & Steps $\downarrow$\\
 \midrule
 GD & Frozen & 42.48 & 32.44 & 20.02 & 44.10 & 41.56 & 19.94\\
GD & AdaJEPA & 72.19 & 55.33 & 14.52 & 58.67 & 54.89 & 17.96\\
GD & \textbf{SE-JEPA} & \textbf{81.59} & \textbf{72.44} & \textbf{11.30} & \textbf{75.68} & \textbf{73.63} & \textbf{13.21}\\
\midrule
CEM & Frozen & 43.81 & 36.22 & 21.08 & 50.86 & 48.33 & 19.56\\
CEM & AdaJEPA & 70.38 & 54.44 & 16.02 & 62.95 & 60.22 & 17.87\\
CEM & \textbf{SE-JEPA} & \textbf{79.08} & \textbf{69.85} & \textbf{13.53} & \textbf{77.05} & \textbf{76.11} & \textbf{14.15}\\
 
 \bottomrule
 \end{tabular}
\end{table*}

\section{Baseline Implementations and Sources}
\label{app:baselines}

\vspace{1mm} \noindent \textbf{Model initialization.}
Frozen and AdaJEPA use the official task-specific checkpoints on PushObj and PushT.
The DINO-WM shape baseline uses the same trajectory count and epoch budget, and its PushT CEM checkpoint is trained for three epochs on 16,000 newly generated random T-shape trajectories.
DreamerV3, IRIS, DINO CLS, and R3M use T-only checkpoints.
DreamerV3 and IRIS serve as frozen offline world models for continuous-action CEM planning, without reward or policy learning.
Configurations and source records are archived with the paper.

\vspace{1mm} \noindent \textbf{Evaluation groups.}
Frozen, AdaJEPA, and \ours{} use the paired held-out targets in \appendixref{app:protocols}.
PushT Shift averages the six shifted conditions, and Color averages the three recolorings; the original condition enters only the seven-condition results.
DreamerV3 and IRIS PushObj results use 50 targets per shape at seed 2000.
For all methods, Unseen averages I, square, and small tee.
Methods within each group share targets and the total action budget.

\vspace{1mm} \noindent \textbf{Published and deformable results.}
CDs for DreamerV3, IRIS, R3M, and DINO CLS are taken from Zhou et al.~\maincite{zhou2024dinowm}{2}, who evaluate ten instances per task.
Our DINO-WM and Frozen deformable rows share the same fixed reference evaluated on 50 instances.

\section{Additional Results}
\label{app:additional}

\subsection{Per-Shape and Per-Condition Results}
\paperxref{Tables}{tab:shape_domains}{} and~\ref{tab:visual_domains} expand the held-out gains in \paperxref{Fig.}{fig:condition_gains}{} into absolute success rates.
On PushObj, the largest gains appear on the unseen shapes; with GD, I rises from 44.7\% to 77.6\% and square from 46.7\% to 61.1\% relative to AdaJEPA.
\ours{} also leads on the four seen shapes in aggregate, reaching 88.4\% (GD) and 86.0\% (CEM) against 84.8\% and 82.3\% for AdaJEPA.
Selective writing leaves familiar dynamics to the fixed pretrained model.
On PushT, \ours{} improves every condition with both planners.
Gains exceed 25 points under red-object and red-goal recoloring, where episodic adaptation must relearn the same appearance correction for every new target.

\begin{table*}[t]
 \centering
 \caption{\textbf{Held-out success (\%) per PushObj shape.} I, square, and small tee are unseen during pretraining; All averages the seven shapes.}
 \label{tab:shape_domains}
 \small
 \begin{tabular}{llcccccccc}
 \toprule
 Planner & Method & I & L & Square & Z & Small tee & Plus & T & All\\
 \midrule
 CEM & Frozen & 32.7 & 52.7 & 23.3 & 41.3 & 52.7 & 43.3 & 60.7 & 43.8\\
CEM & AdaJEPA & 49.3 & 82.0 & 44.7 & 82.0 & 69.3 & 79.3 & 86.0 & 70.4\\
CEM & \textbf{SE-JEPA} & \textbf{72.7} & \textbf{84.2} & \textbf{58.4} & \textbf{86.2} & \textbf{78.4} & \textbf{85.6} & \textbf{88.0} & \textbf{79.1}\\
\midrule
GD & Frozen & 24.0 & 52.7 & 28.0 & 40.0 & 45.3 & 48.7 & 58.7 & 42.5\\
GD & AdaJEPA & 44.7 & \textbf{90.0} & 46.7 & 82.0 & 74.7 & 82.0 & 85.3 & 72.2\\
GD & \textbf{SE-JEPA} & \textbf{77.6} & 88.4 & \textbf{61.1} & \textbf{88.0} & \textbf{78.7} & \textbf{87.6} & \textbf{89.8} & \textbf{81.6}\\
 
 \bottomrule
 \end{tabular}
\end{table*}

\begin{table*}[t]
 \centering
 \caption{\textbf{Held-out success (\%) per PushT condition.} All averages the seven conditions; Shift excludes Original.}
 \label{tab:visual_domains}
 \small
 \setlength{\tabcolsep}{4.5pt}
 \begin{tabular}{llccccccccc}
 \toprule
 Planner & Method & Original & Blur & Noise & Dark & Red agent & Red object & Red goal & All & Shift\\
 \midrule
 CEM & Frozen & 66.0 & 56.0 & 62.7 & 70.7 & 50.0 & 14.0 & 36.7 & 50.9 & 48.3\\
CEM & AdaJEPA & 79.3 & 78.0 & 76.7 & 80.0 & 71.3 & 14.0 & 41.3 & 63.0 & 60.2\\
CEM & \textbf{SE-JEPA} & \textbf{82.7} & \textbf{86.4} & \textbf{86.7} & \textbf{88.0} & \textbf{79.6} & \textbf{47.1} & \textbf{68.9} & \textbf{77.0} & \textbf{76.1}\\
\midrule
GD & Frozen & 59.3 & 48.7 & 55.3 & 58.7 & 50.7 & 6.7 & 29.3 & 44.1 & 41.6\\
GD & AdaJEPA & 81.3 & 77.3 & 73.3 & 73.3 & 58.0 & 10.7 & 36.7 & 58.7 & 54.9\\
GD & \textbf{SE-JEPA} & \textbf{88.0} & \textbf{88.7} & \textbf{84.9} & \textbf{87.3} & \textbf{74.7} & \textbf{44.4} & \textbf{61.8} & \textbf{75.7} & \textbf{73.6}\\
 
 \bottomrule
 \end{tabular}
\end{table*}

\subsection{Deformable Consolidation}
\paperxref{Table}{tab:deform}{} adds standard errors to the deformable columns of \paperxref{Table}{tab:main}{}, and \paperxref{Table}{tab:deform_passes}{} follows consolidation pass by pass.
With the default slow learning rate of $10^{-4}$, CD decreases at every pass on both materials.
A rate of $2\times10^{-4}$ improves rope faster in early passes, but both rates reach similar CD after four passes.

\begin{table}[t]
  \centering
  \caption{\textbf{Deformable manipulation with standard errors.} Mean final CD $\pm$ standard error over 50 instances. \ours{} uses its slow predictor after four passes.}
  \label{tab:deform}
  \small
  \begin{tabular}{lcc}
    \toprule
    Method & Rope CD $\downarrow$ & Granular CD $\downarrow$\\
    \midrule
    DINO-WM & 0.491 $\pm$ 0.059 & 0.301 $\pm$ 0.020\\
LeWM & 0.786 $\pm$ 0.112 & 0.322 $\pm$ 0.018\\
DDP-WM & 0.677 $\pm$ 0.086 & 0.259 $\pm$ 0.012\\
AdaJEPA & 0.629 $\pm$ 0.096 & 0.304 $\pm$ 0.019\\
\textbf{\ours} & \textbf{0.252 $\pm$ 0.045} & \textbf{0.220 $\pm$ 0.012}\\
 
    \bottomrule
  \end{tabular}
\end{table}

\begin{table}[t]
  \centering
  \caption{\textbf{Consolidation pass by pass.} Final CD of \ours{} (mean $\pm$ standard error, 50 instances) at two slow learning rates.}
  \label{tab:deform_passes}
  \footnotesize
  \setlength{\tabcolsep}{5pt}
  \begin{tabular}{ccccc}
    \toprule
    & \multicolumn{2}{c}{lr $10^{-4}$} & \multicolumn{2}{c}{lr $2\times10^{-4}$}\\
    \cmidrule(lr){2-3}\cmidrule(lr){4-5}
    Pass & Rope & Granular & Rope & Granular\\
    \midrule
    1 & 0.469 $\pm$ 0.100 & 0.257 $\pm$ 0.014 & 0.416 $\pm$ 0.070 & 0.260 $\pm$ 0.012\\
2 & 0.342 $\pm$ 0.054 & 0.235 $\pm$ 0.012 & 0.286 $\pm$ 0.050 & \textbf{0.241 $\pm$ 0.012}\\
3 & 0.282 $\pm$ 0.033 & 0.224 $\pm$ 0.010 & 0.283 $\pm$ 0.037 & 0.247 $\pm$ 0.012\\
4 & \textbf{0.252 $\pm$ 0.045} & \textbf{0.220 $\pm$ 0.012} & \textbf{0.243 $\pm$ 0.033} & 0.243 $\pm$ 0.013\\
 
    \bottomrule
  \end{tabular}
\end{table}

\subsection{Shared Memory across Fourteen Domains}
\label{app:mixed14}
This section complements \paperxref{Section}{sec:mixed14}{} and \paperxref{Fig.}{fig:mixed14_learning}{}.
The 14 domains form two recurring families rather than all shape--appearance combinations.
Each round presents the seven visual conditions followed by the seven shapes, with the shape order alternating between rounds.
Initial/goal pairs recur across rounds, while the seeds vary run randomness.
The horizontal baselines in \paperxref{Fig.}{fig:mixed14_learning}{} average archived evaluations, and shading shows the sample SD over six planner-by-chain or planner-by-seed rates.
Its thumbnails show the first episode of each domain from GD chain two in round five, selected by index rather than outcome.

\vspace{1mm} \noindent \textbf{Per-planner results.}
From the first to the fifth round, GD improves from 56.8\% to 69.1\% on visual conditions and from 74.4\% to 81.1\% on shapes; CEM improves from 57.6\% to 68.7\% and from 72.4\% to 82.9\%.
In the fifth round (\paperxref{Table}{tab:mixed14_summary}{}), \ours{} exceeds AdaJEPA by 10.0 (GD) and 10.4 (CEM) percentage points and reduces mean replans by 18.8\% and 17.6\%.
\paperxref{Table}{tab:mixed14_domains}{} shows that \ours{} outperforms AdaJEPA in 12 of 14 domains with GD and 13 with CEM, tying on noise.

\begin{table*}[t]
 \centering
 \caption{\textbf{Mixed 14-domain online results.} Success (\%) and all-episode MPC replans, mean $\pm$ sample SD across three chains (\ours, round five) or three seeds (baselines). Each row has 2,100 episodes, split equally between the two families. Failures use the full budget.}
 \label{tab:mixed14_summary}
 \small
 \begin{tabular}{llcccc}
 \toprule
 Planner & Method & All success $\uparrow$ & Shape success $\uparrow$ & Visual success $\uparrow$ & Replans $\downarrow$\\
 \midrule
 GD & Frozen & 37.7 $\pm$ 0.0 & 42.0 $\pm$ 0.0 & 33.4 $\pm$ 0.0 & 21.29 $\pm$ 0.00\\
GD & AdaJEPA & 65.1 $\pm$ 0.7 & 72.0 $\pm$ 0.9 & 58.2 $\pm$ 0.7 & 16.04 $\pm$ 0.07\\
GD & \textbf{\ours} & \textbf{75.1$\,\pm\,$1.4} & \textbf{81.1$\,\pm\,$2.3} & \textbf{69.1$\,\pm\,$1.6} & \textbf{13.03$\,\pm\,$0.32}\\
\midrule
CEM & Frozen & 38.6 $\pm$ 0.9 & 44.0 $\pm$ 1.0 & 33.2 $\pm$ 2.5 & 21.93 $\pm$ 0.33\\
CEM & AdaJEPA & 65.4 $\pm$ 0.7 & 71.8 $\pm$ 0.4 & 59.0 $\pm$ 1.4 & 17.28 $\pm$ 0.25\\
CEM & \textbf{\ours} & \textbf{75.8$\,\pm\,$0.8} & \textbf{82.9$\,\pm\,$1.0} & \textbf{68.7$\,\pm\,$1.9} & \textbf{14.24$\,\pm\,$0.18}\\
 
 \bottomrule
 \end{tabular}
\end{table*}

\begin{table*}[t]
 \centering
 \caption{\textbf{Mixed-stream success (\%) for every domain.} Each entry averages 150 episodes under the protocol of \paperxref{Table}{tab:mixed14_summary}{}. Original appearance and the T-shape block use different target sets and are retained as separate domains.}
 \label{tab:mixed14_domains}
 \small
 \begin{tabular}{llcccccc}
 \toprule
 & & \multicolumn{3}{c}{GD} & \multicolumn{3}{c}{CEM}\\
 \cmidrule(lr){3-5}\cmidrule(lr){6-8}
 Family & Domain & Frozen & AdaJEPA & \textbf{\ours} & Frozen & AdaJEPA & \textbf{\ours}\\
 \midrule
 Visual & Original & 50.0 & 70.7 & \textbf{78.7} & 46.7 & 72.0 & \textbf{78.0}\\
Visual & Blur & 32.0 & 72.7 & \textbf{78.0} & 30.7 & 70.0 & \textbf{72.7}\\
Visual & Noise & 44.0 & 76.7 & \textbf{78.7} & 41.3 & \textbf{73.3} & \textbf{73.3}\\
Visual & Dark & 48.0 & 72.0 & \textbf{76.0} & 48.7 & 73.3 & \textbf{74.7}\\
Visual & Red agent & 36.0 & 56.0 & \textbf{62.0} & 38.0 & 56.7 & \textbf{71.3}\\
Visual & Red object & 6.0 & 13.3 & \textbf{52.0} & 6.0 & 21.3 & \textbf{50.7}\\
Visual & Red goal & 18.0 & 46.0 & \textbf{58.7} & 21.3 & 46.0 & \textbf{60.0}\\
\midrule
Shape & I & 40.0 & 56.7 & \textbf{82.7} & 26.0 & 54.0 & \textbf{88.0}\\
Shape & L & 52.0 & \textbf{83.3} & 82.0 & 49.3 & 82.7 & \textbf{90.0}\\
Shape & Square & 14.0 & 41.3 & \textbf{60.0} & 26.7 & 38.7 & \textbf{59.3}\\
Shape & Z & 34.0 & 78.7 & \textbf{79.3} & 42.7 & 78.7 & \textbf{83.3}\\
Shape & Small tee & 52.0 & \textbf{78.7} & 76.0 & 56.7 & 80.0 & \textbf{81.3}\\
Shape & Plus & 54.0 & 84.7 & \textbf{94.0} & 46.7 & 81.3 & \textbf{86.7}\\
Shape & T & 48.0 & 80.7 & \textbf{94.0} & 60.0 & 87.3 & \textbf{91.3}\\
 
 \bottomrule
 \end{tabular}
\end{table*}

\subsection{Mixed Rope and Granular Learning}
\label{app:mixed_deformable}
This section details the mixed-material stream in \paperxref{Table}{tab:deform_mixed_main}{}.
A single CEM chain alternates blocks of ten rope and ten granular episodes, five times per round, for three rounds and 300 goals.
No source trajectory is shared across rounds.
Both materials start from the same rope-trained DINO-WM, so granular material is new to the pretrained predictor.
Material labels are used only for analysis; the algorithm receives neither domain identities nor the number of domains.
Besides the relative validation test, a write is accepted only if the episode's prediction error is at most 20 times the candidate predictor's median error on its stored trajectories (\paperxref{Table}{tab:deform_hyperparams}{}).

\vspace{1mm} \noindent \textbf{Domain separation.}
After three rounds, all 80 stored rope episodes and all 150 stored granular episodes reside in their respective material domains (\paperxref{Table}{tab:deform_mixed}{}).
The remaining 70 episodes are 69 familiar rope episodes and one rejected write.
All seven material mismatches among the 660 slow-predictor selections occur in three episodes of the first granular block.
The rope predictor was selected during these replans, yet all three trajectories were stored in the granular domain.

\begin{table}[t]
 \centering
 \caption{\textbf{Mixed deformable learning and domain separation.} One CEM chain, 50 new goal segments per material and round. Stored R/G counts are cumulative rope/granular members. Purity measures the fraction of stored members matching their domain's majority material. Correct retrievals count material-matched slow-predictor selections at MPC replans; selections of the frozen reference are excluded.}
 \label{tab:deform_mixed}
 \small
 \setlength{\tabcolsep}{8pt}
 \adjustbox{max width=\linewidth}{%
 \begin{tabular}{@{}cccccc@{}}
 \toprule
 Round & Rope CD $\downarrow$ & Granular CD $\downarrow$ & Stored R/G & Purity (\%) & Correct retrievals\\
 \midrule
 1 & 0.598 & 0.471 & 27 / 50 & \textbf{100} & 196/203\\
2 & 0.452 & 0.369 & 50 / 100 & \textbf{100} & \textbf{219/219}\\
3 & \textbf{0.412} & \textbf{0.355} & 80 / 150 & \textbf{100} & \textbf{238/238}\\
 
 \bottomrule
 \end{tabular}}
\end{table}

\subsection{Accumulating Fast Updates}
\paperxref{Table}{tab:inheritance}{} retains fast updates as $s_{e+1}=s_e+w\Delta_e$.
Every variant receives the same 100 PushObj goals in five blocks of 20 (I, square, T, I, square).
Full inheritance loses significantly more goals than it recovers relative to reset (two-sided exact McNemar test, $p=0.029$).

Consecutive fast updates also point in nearly unrelated directions.
In the seven-round stream described in \appendixref{sec:selectivity}, the cosine between consecutive fast-update fingerprints has a mean of 0.018 for both planners (\paperxref{Fig.}{fig:directions}{}).
Its absolute value is below 0.1 for 96.3\% of GD pairs and 96.8\% of CEM pairs.
These descriptive statistics come from one sequential stream per planner and support consolidating trajectories rather than accumulating parameter updates.

\begin{figure*}[t]
 \centering
 \includegraphics[width=\textwidth]{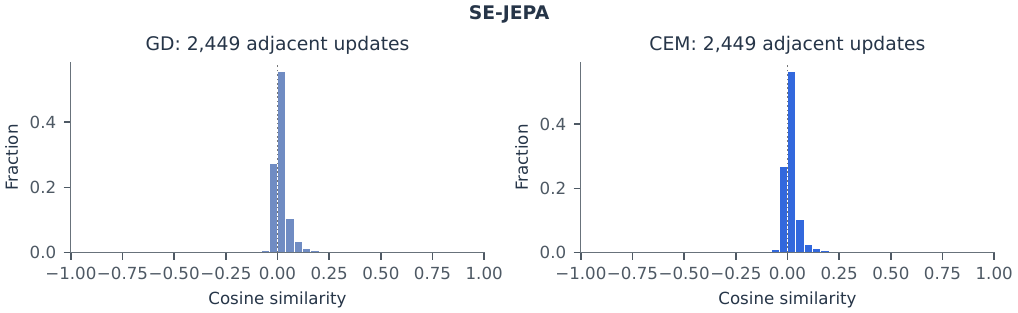}
 \caption{\textbf{Consecutive fast updates are nearly orthogonal.} Cosine similarity between consecutive fast-update fingerprints in the seven-round shape stream (2{,}449 pairs per planner).}
 \label{fig:directions}
\end{figure*}

\subsection{Memory Use in a Seven-Round Stream}
\label{sec:selectivity}
This PushObj stream uses one chain per planner and a slow learning rate of $2\times10^{-4}$.
Each round draws 50 episodes per shape without replacement, excluding held-out sources, for 2{,}450 episodes per planner.
Online success rises from 70.6\% to 78.6\% with GD and from 72.3\% to 78.3\% with CEM between the first and seventh rounds.
Memory use is selective: only 25\% (GD) and 27\% (CEM) of episodes are written, while the rest are familiar, pending, or rejected by validation.
About one quarter of episodes retrieve a slow predictor; the others adapt from the pretrained model.

\subsection{Component Ablations}
\label{app:component_ablation}
\paperxref{Tables}{tab:gd_components_full}{} and~\ref{tab:planner_components_full} extend \paperxref{Table}{tab:component_ablation}{} with mean replans and additional memory designs.
Each variant learns for five rounds in one stream per task and planner, and its final memory is tested with writes disabled on seeds 2000, 2100, and 2200 (50 goals per domain and seed, 1,050 per cell).
All variants within a task share the pretrained initialization.

\vspace{1mm} \noindent \textbf{Memory design.}
Episodic reset removes persistent memory, and direct inheritance carries fast weights to the next goal.
Two shared-predictor variants either train one slow predictor by replay or average fast parameters with an EMA ($\alpha=10^{-4}$).
Under GD (\paperxref{Table}{tab:gd_components_full}{}), replay outperforms EMA on both tasks, and domain-specific replay achieves the highest success.
Relative to direct inheritance, \ours{} reduces mean replans from 16.12 to 11.37 on PushObj and from 22.33 to 12.93 on PushT.
The last two rows disable familiarity protection separately for retrieval and for memory writes.
On PushObj, disabling familiar-read protection did not lead to any retrieval of a slow predictor during evaluation. Both this variant and episodic reset therefore used the pretrained predictor \(P_0\) as the base for fast adaptation within each episode, yielding identical success rates and mean replan counts.

\begin{table}[!ht]
 \centering
 \caption{\textbf{Memory-design ablations (GD).} Each entry reports success (\%) / all-episode MPC replans on 1,050 held-out goals. Variants share five learning rounds and paired test goals.}
 \label{tab:gd_components_full}
 \small
 \setlength{\tabcolsep}{8pt}
 \begin{tabular}{@{}lcc@{}}
 \toprule
 Variant & PushObj & PushT\\
 \midrule
 \textbf{Full \ours} & \textbf{81.43} / \textbf{11.37} & \textbf{76.57} / \textbf{12.93}\\
Episodic reset & 76.95 / 12.54 & 65.90 / 15.47\\
Inherit fast weights & 64.67 / 16.12 & 40.19 / 22.33\\
Shared predictor: EMA & 75.71 / 12.84 & 66.29 / 15.30\\
Shared predictor: replay & 81.24 / 11.41 & 75.14 / 13.26\\
No familiar-read protection & 76.95 / 12.54 & 76.19 / 13.16\\
No familiar-write protection & 79.43 / 11.76 & 73.81 / 13.66\\
 
 \bottomrule
 \end{tabular}
\end{table}

\vspace{1mm} \noindent \textbf{Gates and learning rates.}
Removing maturity screening lets immature predictors compete for retrieval, while prediction-error selection is kept.
Removing familiarity protection lets familiar episodes retrieve slow predictors and enter replay.
Familiarity thus controls which episodes participate in memory, whereas maturity controls when a predictor is ready for reuse.
In \paperxref{Table}{tab:planner_components_full}{}, mean replans largely follow the success ordering, so the conclusions in the main text also hold for interaction cost.

\begin{table*}[t]
 \centering
 \caption{\textbf{Gate and slow learning-rate ablations (GD and CEM).} Each entry is success (\%) / mean MPC replans over 1,050 held-out goals. The final state of one complete five-round learning stream is evaluated for each task and planner.}
 \label{tab:planner_components_full}
 \small
 \setlength{\tabcolsep}{8pt}
 \begin{tabular}{@{}lcccc@{}}
 \toprule
 & \multicolumn{2}{c}{PushObj} & \multicolumn{2}{c}{PushT}\\
 \cmidrule(lr){2-3}\cmidrule(lr){4-5}
 Variant & GD & CEM & GD & CEM\\
 \midrule
 \textbf{Full \ours} & 81.43 / 11.37 & 78.57 / 13.64 & 76.57 / 12.93 & \textbf{77.90} / \textbf{14.02}\\
No maturity screening & 80.86 / 11.46 & 79.24 / 13.51 & \textbf{78.10} / \textbf{12.63} & 75.05 / 14.58\\
No familiarity protection & \textbf{83.14} / \textbf{10.90} & \textbf{79.62} / \textbf{13.12} & 73.14 / 13.47 & 75.14 / 14.18\\
\midrule
\multicolumn{5}{@{}l}{\textbf{Full \ours: fixed slow learning rate}}\\
$5\times10^{-5}$ & 79.62 / 11.79 & \textbf{79.14} / \textbf{13.56} & 73.05 / 13.65 & 75.81 / 14.31\\
$10^{-4}$ (default) & 81.43 / 11.37 & 78.57 / 13.64 & 76.57 / 12.93 & \textbf{77.90} / \textbf{14.02}\\
$2\times10^{-4}$ & \textbf{82.38} / \textbf{11.26} & 78.10 / 13.89 & 76.86 / 12.86 & 76.76 / 14.33\\
$3\times10^{-4}$ & 80.57 / 11.72 & 77.14 / 14.01 & \textbf{78.10} / \textbf{12.78} & 75.05 / 14.45\\
$5\times10^{-4}$ & 78.38 / 12.40 & 74.38 / 14.54 & 74.38 / 13.53 & 73.52 / 15.08\\
 
 \bottomrule
 \end{tabular}
\end{table*}

\subsection{Predictor Selection on Unseen Trajectories}
\label{app:mature_routing}
\paperxref{Table}{tab:mature_routing}{} evaluates the three mature predictors retained after five CEM learning rounds on PushT.
We draw 50 new start/goal pairs, each rendered under seven visual conditions, and a disjoint set of ten calibration pairs, all excluded from the preceding learning and evaluation streams.
The pretrained predictor collects one CEM trajectory per pair, and all candidates are scored on the same states and actions with learning disabled.

\vspace{1mm} \noindent \textbf{Selection protocol.}
The deployed rule scores the pretrained reference and eligible mature predictors on the first three transitions.
Familiarity protection retains the reference in 195 of the 350 task--condition pairs, and the remaining 155 admit all four candidates.
The fixed-domain comparator pairs the reference with one mature predictor, chosen from the 70 calibration evaluations before testing.
Random selection averages uniformly over eligible candidates, and the oracle takes the lowest future error among all four.
One-step MSE is measured strictly after the selection prefix.
Five-step MSE rolls out autoregressively with the recorded actions and requires a full five-step window, which 266 pairs provide.
All methods use the same pairs, and the seven conditions are weighted equally.

\vspace{1mm} \noindent \textbf{Uncertainty.}
In units of $10^{-3}$, selection reduces one-step MSE by 9.09 [5.89, 12.85] relative to the fixed-domain comparator and by 33.71 [28.47, 39.32] relative to random selection.
Brackets give paired 95\% intervals from 2,000 bootstrap resamples over the 50 physical tasks, keeping all visual conditions of a task together.
Among pairs eligible for retrieval, the selected predictor matches the future-best predictor in 82.7\% of cases, averaged over conditions.

\subsection{Learning a Single Unseen Shape}
\label{app:i_only}
This section expands \paperxref{Table}{tab:i_only}{}.
A single GD stream learns only the unseen I shape for seven rounds of 50 episodes.
After each round, persistent memory is fixed and evaluated with within-episode adaptation on three test seeds with 50 goals per shape and seed.
These 1,050 evaluations cover 950 distinct initial/goal pairs and are paired across rounds and methods.
Episodic reset is a constant reference.

\vspace{1mm} \noindent \textbf{Specialization and retention.}
Most of the gain on I appears after the first round, which raises success to 66.0\% (\paperxref{Table}{tab:i_only_rounds}{}).
Success on the other six shapes stays between 80.56\% and 81.78\% throughout.
After the final round, L, T, Z, plus, small tee, and square reach 88.67\%, 92.00\%, 88.67\%, 88.00\%, 80.00\%, and 53.33\% with \ours{}, compared with 48.00\%, 38.00\%, 28.00\%, 29.33\%, 38.00\%, and 21.33\% under fast-weight inheritance.
Accumulating local updates therefore both limits specialization and transfers negatively to other shapes.

\vspace{1mm} \noindent \textbf{Memory use.}
A mature slow predictor is available after every round and is retrieved in 20.1--23.1\% of held-out episodes, including 211 of 1,050 after round seven.
The final memory holds 327 trajectories after 5,085 slow optimization steps.

\begin{table}[!ht]
\centering
\caption{\textbf{Success throughout I-only learning.} Each entry is a percentage over the fixed evaluation set. The other-six column averages the shapes absent from this learning stream.}
\label{tab:i_only_rounds}
\small
\begin{tabular}{@{}lcccc@{}}
\toprule
 & \multicolumn{2}{c}{Fast inheritance} & \multicolumn{2}{c}{\textbf{\ours}}\\
\cmidrule(lr){2-3}\cmidrule(lr){4-5}
Round & I & Other six & I & Other six\\
\midrule
1 & 38.00 & 51.89 & 66.00 & 81.56\\
2 & 37.33 & 48.78 & 68.67 & 80.78\\
3 & \textbf{47.33} & \textbf{54.67} & 73.33 & 81.56\\
4 & 39.33 & 51.67 & 74.00 & 81.33\\
5 & 35.33 & 45.22 & 74.67 & 80.89\\
6 & 45.33 & 44.67 & \textbf{77.33} & 80.56\\
7 & 38.00 & 33.78 & 76.67 & \textbf{81.78}\\
 
\bottomrule
\end{tabular}
\end{table}

\FloatBarrier
\section{Qualitative Examples and Reproducibility}
\label{app:qualitative}

\subsection{Planning Sequences on PushObj and PushT}
\vspace{1mm} \noindent \textbf{Provenance of \paperxref{Fig.}{fig:qualitative}{}.}
Both GD examples were selected because Frozen, DINO-WM, and AdaJEPA fail while \ours{} succeeds; aggregate success always uses the complete evaluation sets.
The unseen-I example comes from the paired shape evaluation, and its DINO-WM frames replay recorded actions.
The same replay procedure exactly reproduces all 31 archived Frozen frames and states.
The red-object example comes from round five of the 14-domain stream (\appendixref{app:mixed14}).
Its DINO-WM final frame comes from the 30-replan simulation, and the other final frames replay the recorded actions in the original simulator.

\begin{figure*}[!t]
 \centering
 \includegraphics[width=\textwidth]{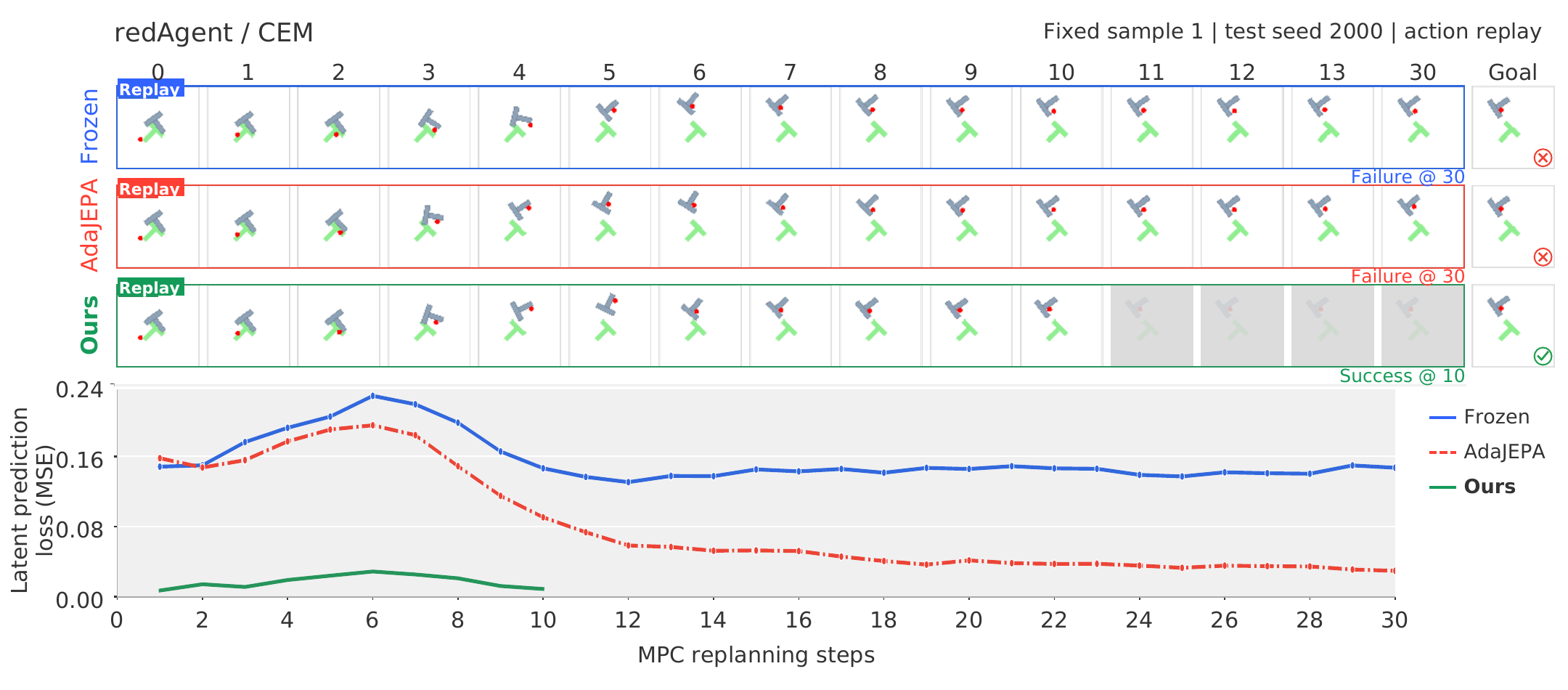}
 \caption{\textbf{Planning under the red-agent shift with CEM.} Layout follows \paperxref{Fig.}{fig:full_shape}{}, with frames replayed from recorded actions. \ours{} reaches the goal at replan 10, whereas Frozen and AdaJEPA exhaust all 30 replans. AdaJEPA lowers its prediction loss but still misses the goal.}
 \label{fig:full_visual}
\end{figure*}

\vspace{1mm} \noindent \textbf{Complete CEM sequences.}
\paperxref{Figs.}{fig:full_shape}{} and~\ref{fig:full_visual} show the full execution strips, termination markers, and prediction-loss curves for an unseen-square and a red-agent episode.
Both use the first held-out target of test seed 2000 from the original execution archive.
Replay marks frames reconstructed in the original simulator from recorded actions.
On the square, \ours{} reaches the goal three replans earlier than AdaJEPA, while Frozen fails.
On the red-agent shift, \ours{} succeeds where both baselines fail.
These cases illustrate two benefits of memory: reaching a shared goal sooner and solving goals that episode-local adaptation misses.

\begin{figure*}[!t]
 \centering
 \includegraphics[width=\textwidth]{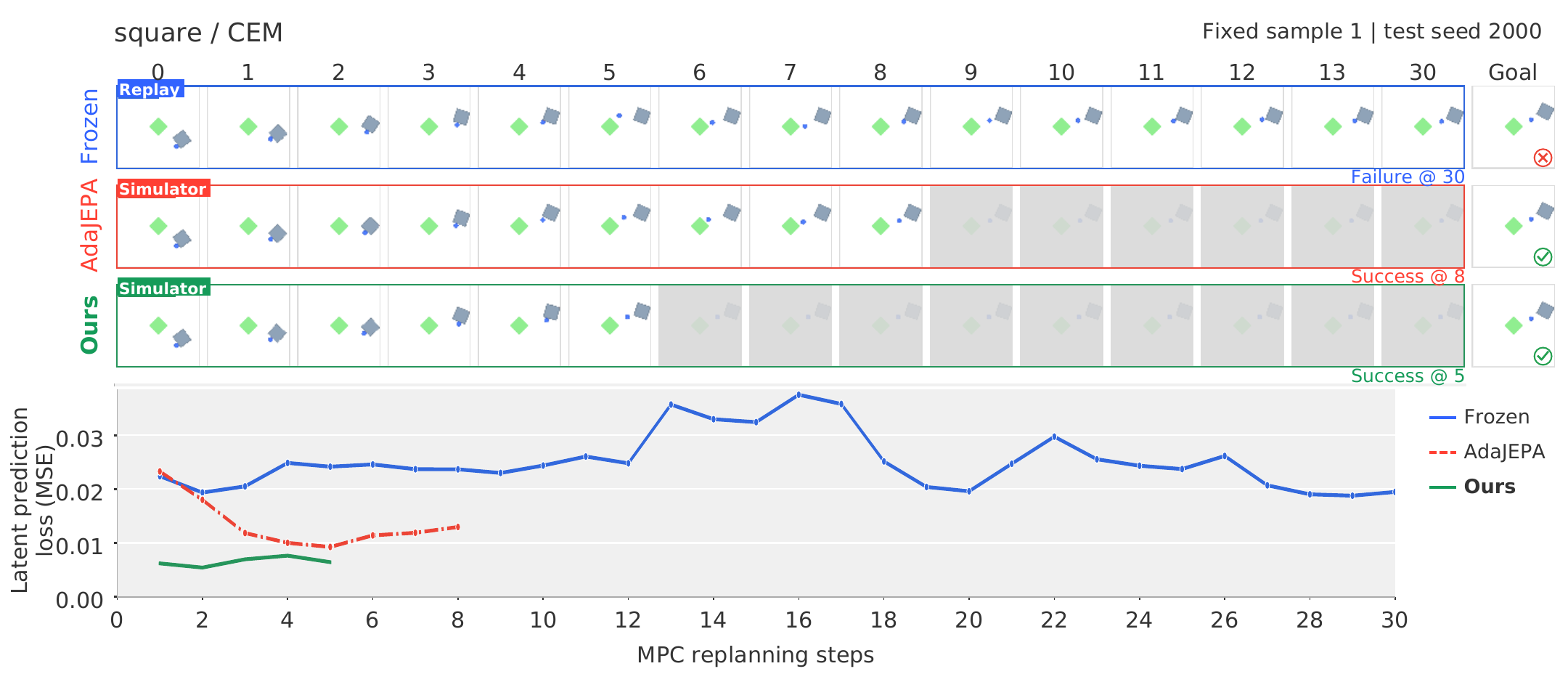}
 \caption{\textbf{Planning on the unseen square shape under CEM.} Frames show replans 0--13 and 30, followed by the goal; gray frames hold the terminal state. \ours{} succeeds at replan 5 and AdaJEPA at replan 8, whereas Frozen exhausts all 30 replans. Curves show each method's prediction MSE over up to five recent transitions and end at termination.}
 \label{fig:full_shape}
\end{figure*}

The plotted loss is the latent prediction MSE over up to five recent transitions on visual and proprioceptive features, excluding action dimensions.
Each curve uses its own method's weights, so absolute loss levels are not comparable across methods, and task success remains the common performance measure.
On the red-agent episode, for example, AdaJEPA reduces its prediction loss yet misses the goal.

\subsection{Deformable Rollout Sequences}
\paperxref{Fig.}{fig:deform_reference}{} shows complete sequences for rope example 1 and granular example 2 under the same 15-push budget.
Both come from the repeated 50-instance stream and were selected because the consolidated predictor ends with lower CD than both baselines.
They use the third-pass slow predictor without within-episode adaptation, whereas \paperxref{Table}{tab:main}{} reports fourth-pass aggregates.
DINO-WM frames replay recorded actions in FleX, AdaJEPA and \ours{} frames were saved during planning, and final CDs come from run logs.
On rope, both baselines reduce CD from 1.72 only to 1.54, whereas \ours{} reaches 0.31.
On granular material, \ours{} reaches 0.14 compared with 0.33 for both baselines.

\begin{figure*}[p]
 \centering
 \includegraphics[width=\textwidth]{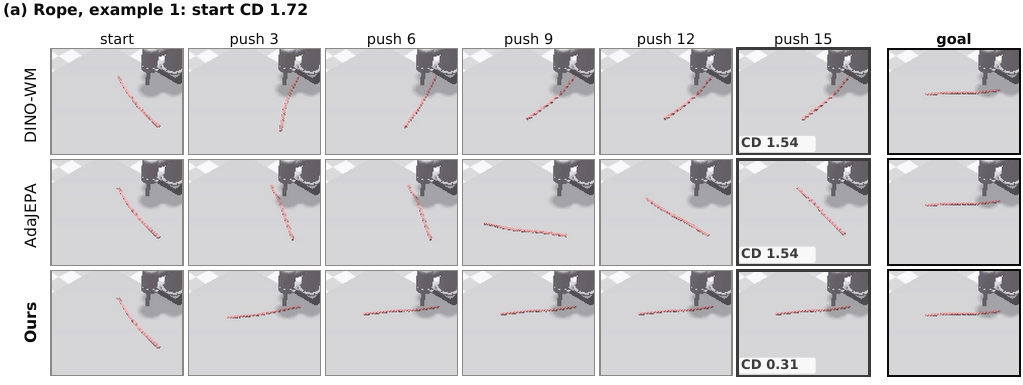}
 \par\medskip
 \includegraphics[width=\textwidth]{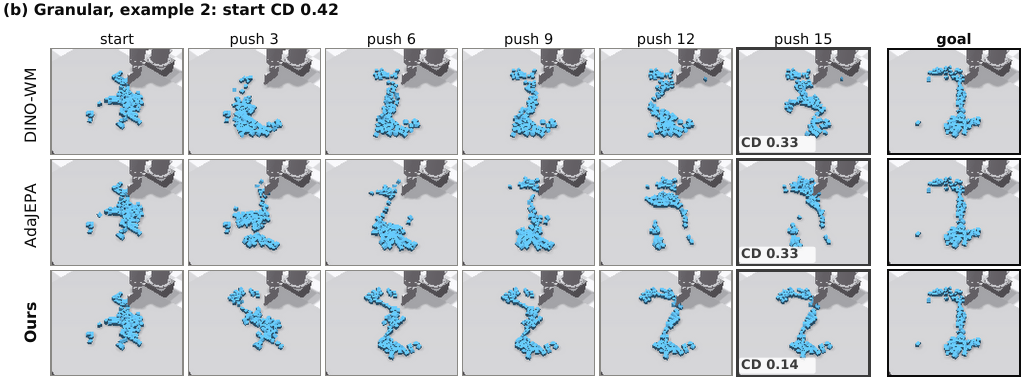}
 \caption{\textbf{Rope and granular execution sequences.} Top: rope example 1. Bottom: granular example 2. All methods share the initial configuration and goal. Columns show pushes 0, 3, 6, 9, 12 and 15, then the goal. Final-frame labels report CD (lower is better).}
 \label{fig:deform_reference}
\end{figure*}

\end{document}